\documentclass[11pt,a4paper]{article}

\usepackage[hyperref]{conll2018}

\usepackage[colorinlistoftodos]{todonotes}

\usepackage{times}
\usepackage{latexsym}

\ifxetex
    \usepackage{fontspec}
\else
    \usepackage[T1]{fontenc}
    \usepackage[utf8]{inputenc}
\fi

\usepackage{graphicx}

\usepackage{rotating}

\usepackage{amsmath}
\usepackage{amsfonts}
\usepackage{amssymb}
\usepackage{amsthm}

\usepackage{algorithm}
\usepackage{algorithmic}

\usepackage{makecell}
\usepackage{lipsum}

\theoremstyle{definition}
\newtheorem{examplex}{Example}

\newenvironment{example}
  {\pushQED{\qed}\examplex}
  {\popQED\endexamplex}

\usepackage{mathtools}
\usepackage{xcolor}

\usepackage{booktabs}
\usepackage{comment}

\usepackage[skip=5pt]{caption}

\usepackage[normalem]{ulem}

\usepackage{paralist}
\usepackage{multirow}
\usepackage{array}
\usepackage{tabularx}
\usepackage{adjustbox}

\newcommand{\cfbox}[2]{%
    \colorlet{currentcolor}{.}%
    {\color{#1}%
    \fbox{\color{currentcolor}#2}}%
}

\newcolumntype{L}[1]{>{\raggedright\let\newline\\\arraybackslash\hspace{0pt}}m{#1}}
\newcolumntype{C}[1]{>{\centering\let\newline\\\arraybackslash\hspace{0pt}}m{#1}}
\newcolumntype{R}[1]{>{\raggedleft\let\newline\\\arraybackslash\hspace{0pt}}m{#1}}

\usepackage{diagbox}

\DeclareMathOperator{\softmax}{softmax}
\DeclareMathOperator{\score}{score}

\newcommand{\sspace}{\ensuremath{\mathcal{S}}}
\newcommand{\sset}{\ensuremath{S}}
\newcommand{\var}{\ensuremath{X}}
\newcommand{\rel}[1]{{\ensuremath{\mathrm{#1}}}}

\newcommand{\xrule}[1]{\ensuremath{\mathbf{R_{#1}}}}

\newcommand{\aset}[2]{ \mathcal{A}_{\text{#1}}^{#2} }

\newcommand{\bfa}{\ensuremath{\mathbf{a}}}
\newcommand{\bfb}{\ensuremath{\mathbf{b}}}

\newcommand{\bfx}{\ensuremath{\mathbf{x}}}

\newcommand{\DAM}{\textrm{DAM}}
\newcommand{\ESIM}{\textrm{ESIM}}
\newcommand{\cBiLSTM}{\textrm{cBiLSTM}}

\newcommand{\card}[1]{\ensuremath{\left\vert{#1}\right\vert}}

\usepackage{xspace}

\newcommand*{\ie}{i.e.\@\xspace}

\makeatletter
\newcommand*{\etc}{%
	\@ifnextchar{.}%
	{etc}%
	{etc.\@\xspace}%
}
\makeatother

\newcommand{\Real}{\ensuremath{\mathbb{R}}}
\newcommand{\Natural}{\ensuremath{\mathbb{N}}}

\newcommand{\data}{\ensuremath{ \mathcal{D} }}

\newcommand{\loss}{\ensuremath{  \mathcal{J} }}
\newcommand{\dloss}{\ensuremath{ \loss_{\mathcal{D}} }}
\newcommand{\iloss}{\ensuremath{ \loss_{\mathcal{I}} }}

\usepackage{bbm}
\newcommand{\ind}{\ensuremath{\mathbbm{1}}}
\newcommand{\params}{\ensuremath{\Theta}}

\newcommand{\langm}{\ensuremath{p_{\mathcal{L}}}}

\usepackage[capitalise]{cleveref}

\aclfinalcopy

\title{Adversarially Regularising Neural NLI Models\\to Integrate Logical Background Knowledge}

\author{Pasquale Minervini \\
  University College London \\
  {\tt p.minervini@cs.ucl.ac.uk} \\\And
  Sebastian Riedel \\
  University College London \\
  {\tt s.riedel@cs.ucl.ac.uk} \\}

\date{}

\begin{document}

\maketitle

\begin{abstract}
Adversarial examples are inputs to machine learning models designed to cause the model to make a mistake.
They are useful for understanding the shortcomings of machine learning models, interpreting their results, and for regularisation.
In NLP, however, most example generation strategies produce input text by using known, pre-specified semantic transformations, requiring significant manual effort and in-depth understanding of the problem and domain.
In this paper, we investigate the problem of automatically generating adversarial examples that violate a set of given First-Order Logic constraints in Natural Language Inference (NLI).
We reduce the problem of identifying such adversarial examples to a combinatorial optimisation problem, by maximising a quantity measuring the degree of violation of such constraints and by using a language model for generating linguistically-plausible examples.
Furthermore, we propose a method for adversarially regularising neural NLI models for incorporating background knowledge.
Our results show that, while the proposed method does not always improve results on the SNLI and MultiNLI datasets, it significantly and consistently increases the predictive accuracy on adversarially-crafted datasets -- up to a 79.6\% relative improvement -- while drastically reducing the number of background knowledge violations.
Furthermore, we show that adversarial examples \emph{transfer} among model architectures, and that the proposed adversarial training procedure improves the robustness of NLI models to adversarial examples.
\end{abstract}
\section{Introduction} \label{sec:introduction}
An open problem in Artificial Intelligence is quantifying the extent to which algorithms exhibit intelligent behaviour~\citep{DBLP:journals/ai/Levesque14}.
In Machine Learning, a standard procedure consists in estimating the \emph{generalisation error}, \ie the prediction error over an independent test sample~\citep{hastie01statisticallearning}.
However, machine learning models can succeed simply by recognising patterns that happen to be predictive on instances in the test sample, while ignoring deeper phenomena~\citep{DBLP:journals/jbi/RimellC09,DBLP:conf/acl/PapernoKLPBPBBF16}.
Adversarial examples are inputs to machine learning models designed to cause the model to make a mistake~\citep{42503,DBLP:journals/corr/GoodfellowSS14}.
In Natural Language Processing (NLP) and Machine Reading, generating adversarial examples can be really useful for understanding the shortcomings of NLP models~\citep{DBLP:conf/emnlp/JiaL17,DBLP:journals/corr/KannanV17} and for regularisation~\citep{DBLP:conf/uai/MinerviniDRR17}.
In this paper, we focus on the problem of generating adversarial examples for Natural Language Inference (NLI) models in order to gain insights about the inner workings of such systems, and regularising them.
NLI, also referred to as Recognising Textual Entailment~\citep{Fyodorov-etal:2000,Condoravdi-etAl:2003,DBLP:conf/mlcw/DaganGM05}, is a central problem in language understanding~\citep{katz1972semantic,DBLP:conf/naacl/BosM05,vanBenthem08NATLOG,maccartney2009extended}, and thus it is especially well suited to serve as a benchmark task for research in machine reading.
In NLI, a model is presented with two sentences, a \emph{premise} $p$ and a \emph{hypothesis} $h$, and  the goal is to determine whether $p$ semantically entails
$h$.
The problem of acquiring large amounts of labelled data for NLI was addressed with the creation of the SNLI~\citep{DBLP:conf/emnlp/BowmanAPM15} and MultiNLI~\citep{DBLP:journals/corr/WilliamsNB17} datasets.
In these processes, annotators were presented with a \emph{premise} $p$ drawn from a corpus, and were required to generate three new sentences (\emph{hypotheses}) based on $p$, according to the following criteria:
\begin{inparaenum}[\itshape a)\upshape]
 \item Entailment -- $h$ is definitely true given $p$ ($p$ entails $h$);
 \item Contradiction -- $h$ is definitely not true given $p$ ($p$ contradicts $h$); and
 \item Neutral -- $h$ might be true given $p$.
\end{inparaenum}
Given a premise-hypothesis sentence pair $(p, h)$, a NLI model is asked to classify the relationship between $p$ and $h$ -- \ie either \emph{entailment}, \emph{contradiction}, or \emph{neutral}.
Solving NLI requires to fully capture the sentence meaning by handling complex linguistic phenomena like lexical entailment, quantification, co-reference, tense, belief, modality, and lexical and syntactic ambiguities~\citep{DBLP:journals/corr/WilliamsNB17}.
In this work, we use adversarial examples for:
\begin{inparaenum}[\itshape a)\upshape]
 \item identifying cases where models violate existing background knowledge, expressed in the form of \emph{logic rules}, and
 \item training models that are \emph{robust} to such violations.
\end{inparaenum}
The underlying idea in our work is that NLI models should adhere to a set of structural constraints that are intrinsic to the human reasoning process.
For instance, \emph{contradiction} is inherently \emph{symmetric}: if a sentence $p$ contradicts a sentence $h$, then $h$ contradicts $p$ as well.
Similarly, entailment is both \emph{reflexive} and \emph{transitive}.
It is reflexive since a sentence $a$ is always entailed by (\ie is true given) $a$.
It is also transitive, since if $a$ is entailed by $b$, and $b$ is entailed by $c$, then $a$ is entailed by $c$ as well.
\begin{example}[Inconsistency] \label{ex:example}
Consider three sentences $a$, $b$ and $c$ each describing a situation, such as:
\begin{inparaenum}[\itshape a)\upshape]
 \item ``The girl plays'',
 \item ``The girl plays with a ball'', and
 \item ``The girl plays with a red ball''.
\end{inparaenum}
Note that if $a$ is entailed by $b$, and $b$ is entailed by $c$, then also $a$ is entailed by $c$.
If a NLI model detects that $b$ entails $a$, $c$ entails $b$, but $c$ does not entail $a$, we know that it is making an error (since its results are inconsistent), even though we may not be aware of the sentences $a$, $b$, and $c$ and the true semantic relationships holding between them.
\end{example}
Our adversarial examples are different from those used in other fields such as computer vision, where they typically consist in small, semantically invariant perturbations that result in drastic changes in the model predictions.
In this paper, we propose a method for generating adversarial examples that cause a model to violate pre-existing background knowledge (\cref{sec:generating}), based on reducing the generation problem to a combinatorial optimisation problem.
Furthermore, we outline a method for incorporating such background knowledge into models by means of an \emph{adversarial training} procedure (\cref{sec:regularisation}).
Our results (\cref{sec:experiments}) show that, even though the proposed adversarial training procedure does not sensibly improve accuracy on SNLI and MultiNLI, it yields significant relative improvement in accuracy (up to 79.6\%) on adversarial datasets.
Furthermore, we show that adversarial examples \emph{transfer} across models, and that the proposed method allows training significantly more robust NLI models.
\section{Background} \label{sec:basics}
\paragraph{Neural NLI Models.}
In NLI, in particular on the Stanford Natural Language Inference (SNLI)~\citep{DBLP:conf/emnlp/BowmanAPM15} and MultiNLI~\citep{DBLP:journals/corr/WilliamsNB17} datasets, neural NLI models -- end-to-end differentiable models that can be trained via gradient-based optimisation -- proved to be very successful, achieving state-of-the-art results~\citep{rocktaschel2016reasoning,DBLP:conf/emnlp/ParikhT0U16,DBLP:conf/acl/ChenZLWJI17}.
Let $\sspace$ denote the set of all possible sentences, and let $a = (a_{1}, \ldots, a_{\ell_{a}}) \in \sspace$ and $b = (b_{1}, \ldots, b_{\ell_{b}}) \in \sspace$ denote two input sentences -- representing the premise and the hypothesis -- of length $\ell_{a}$ and $\ell_{b}$, respectively.
In neural NLI models, all words $a_{i}$ and $b_{j}$ are typically represented by $k$-dimensional \emph{embedding vectors} $\bfa_{i}, \bfb_{j} \in \Real^{k}$.
As such, the sentences $a$ and $b$ can be encoded by the sentence \emph{embedding matrices} $\bfa \in \Real^{k \times \ell_{a}}$ and $\bfb \in \Real^{k \times \ell_{b}}$, where the columns $\bfa_{i}$ and $\bfb_{j}$ respectively denote the embeddings of words $a_{i}$ and $b_{j}$.
Given two sentences $a, b \in \sspace$, the goal of a NLI model is to identify the semantic relation between $a$ and $b$, which can be either \emph{entailment}, \emph{contradiction}, or \emph{neutral}.
For this reason, given an instance, neural NLI models compute the following conditional probability distribution over all three classes:
\begin{equation} \label{eq:nnli}
 \begin{aligned}
  p_{\params}({}\cdot{} \mid a, b) & = & & \softmax(\score_{\params}(\bfa, \bfb))
 \end{aligned}
\end{equation}
\noindent where $\score_{\params} : \Real^{k \times \ell_{a}} \times \Real^{k \times \ell_{b}} \rightarrow \Real^{3}$ is a model-dependent \emph{scoring function} with parameters $\params$, and $\softmax(\bfx)_{i} = \exp\{x_i\}/\sum_j \exp\{x_j\}$ denotes the softmax function.
Several scoring functions have been proposed in the literature, such as the conditional Bidirectional LSTM (\cBiLSTM)~\citep{rocktaschel2016reasoning}, the Decomposable Attention Model (\DAM{})~\citep{DBLP:conf/emnlp/ParikhT0U16}, and the Enhanced LSTM model (\ESIM{})~\citep{DBLP:conf/acl/ChenZLWJI17}.
One desirable quality of the scoring function $\score_{\params}$ is that it should be \emph{differentiable} with respect to the model parameters $\params$, which allows the neural NLI model to be trained from data via back-propagation.
\paragraph{Model Training.}
Let $\data = $ $\{ (x_{1}, y_{1}),$ $\ldots,$ $(x_{m}, y_{m})\}$ represent a NLI dataset, where $x_{i}$ denotes the $i$-th premise-hypothesis sentence pair, and $y_{i} \in \{ 1, \ldots, K \}$ their relationship, where $K \in \Natural$ is the number of possible relationships -- in the case of NLI, $K = 3$.
The model is trained by minimising a \emph{cross-entropy loss} $\dloss$ on $\data$:
\begin{equation} \label{eq:loss}
    \dloss(\data, \params) = - \sum_{i = 1}^{m} \sum_{k = 1}^{K} \ind \{ y_{i} = k \} \log(\hat{y}_{i, k})
\end{equation}
\noindent where $\hat{y}_{i, k} = p_{\params}(y_{i} = k \mid x_{i})$ denotes the probability of class $k$ on the instance $x_{i}$ inferred by the neural NLI model as in \cref{eq:nnli}.
In the following, we analyse the behaviour of neural NLI models by means of \emph{adversarial examples} -- inputs to machine learning models designed to cause the model to commit mistakes.
In computer vision models, adversarial examples are created by adding a very small amount of noise to the input~\citep{42503,DBLP:journals/corr/GoodfellowSS14}: these perturbations do not change the semantics of the images, but they can drastically change the predictions of computer vision models.
In our setting, we define an adversary whose goal is finding \emph{sets} of NLI instances where the model fails to be consistent with available background knowledge, encoded in the form of First-Order Logic (FOL) rules.
In the following sections, we define the corresponding optimisation problem, and propose an efficient solution.
\section{Background Knowledge} \label{sec:background}
For analysing the behaviour of NLI models, we verify whether they agree with the provided background knowledge, encoded by a set of FOL rules.
Note that the three NLI classes --  \emph{entailment}, \emph{contradiction}, and \emph{neutrality} -- can be seen as \emph{binary logic predicates}, and we can define FOL formulas for describing the formal relationships that hold between them.
In the following, we denote the predicates associated with entailment, contradiction, and neutrality as $\rel{ent}$, $\rel{con}$, and $\rel{neu}$, respectively.
By doing so, we can represent semantic relationships between sentences via logic atoms.
For instance, given three sentences $s_{1}, s_{2}, s_{3} \in \sspace$, we can represent the fact that $s_{1}$ entails $s_{2}$ and $s_{2}$ contradicts $s_{3}$ by using the logic atoms $\rel{ent}(s_{1}, s_{2})$ and $\rel{con}(s_{2}, s_{3})$.
Let $X_{1}, \ldots, X_{n}$ be a set of universally quantified variables.
We define our background knowledge as a set of FOL rules, each having the following $\rel{body} \Rightarrow \rel{head}$ form:
\begin{equation} \label{eq:rule}
 \rel{body}(X_{1}, \ldots, X_{n}) \Rightarrow \rel{head}(X_{1}, \ldots, X_{n}),
\end{equation}
\noindent where $\rel{body}$ and $\rel{head}$ represent the \emph{premise} and the \emph{conclusion} of the rule -- if $\rel{body}$ holds, $\rel{head}$ holds as well.
In the following, we consider the rules $\xrule{1}, \ldots, \xrule{5}$ outlined in \cref{tab:rules}.
Rule $\xrule{1}$ enforces the constraint that entailment is reflexive;
rule $\xrule{2}$ that contradiction should always be symmetric (if $s_{1}$ contradicts $s_{2}$, then $s_{2}$ contradicts $s_{1}$ as well);
rule $\xrule{5}$ that entailment is transitive;
while rules $\xrule{3}$ and $\xrule{4}$ describe the formal relationships between the \emph{entailment}, \emph{neutral}, and \emph{contradiction} relations.
In \cref{sec:generating} we propose a method to automatically generate sets of sentences that violate the rules outlined in \cref{tab:rules} -- effectively generating \emph{adversarial examples}.
Then, in \cref{sec:regularisation} we show how we can leverage such adversarial examples by generating them on-the-fly during training and using them for regularising the model parameters, in an \emph{adversarial training} regime.
\section{Generating Adversarial Examples} \label{sec:generating}
\begin{table}[t]
\centering
\resizebox{\columnwidth}{!}{%
 \begin{tabular}{cl}
  \toprule
   \multicolumn{2}{c}{\bf NLI Rules} \\
  \midrule
   $\xrule{1}$ & $\top \Rightarrow \rel{ent}(\var_{1}, \var_{1})$ \\
   $\xrule{2}$ & $\rel{con}(\var_{1}, \var_{2}) \Rightarrow \rel{con}(\var_{2}, \var_{1})$ \\
   $\xrule{3}$ & $\rel{ent}(\var_{1}, \var_{2}) \Rightarrow \lnot \rel{con}(\var_{2}, \var_{1})$ \\
   $\xrule{4}$ & $\rel{neu}(\var_{1}, \var_{2}) \Rightarrow \lnot \rel{con}(\var_{2}, \var_{1})$ \\
   $\xrule{5}$ & $\rel{ent}(\var_{1}, \var_{2}) \land \rel{ent}(\var_{2}, \var_{3}) \Rightarrow \rel{ent}(\var_{1}, \var_{3})$ \\
  \bottomrule
 \end{tabular}
}%
\caption{First-Order Logic rules defining desired properties of NLI models: $X_{i}$ are universally quantified variables, and operators $\land$, $\lnot$, and $\top$ denote logic conjunction, negation, and tautology.}
\label{tab:rules}
\end{table}
In this section, we propose a method for efficiently \emph{generating} adversarial examples for NLI models -- \ie examples that make the model violate the background knowledge outlined in \cref{sec:background}.
\subsection{Inconsistency Loss} \label{ssec:inconsistency}
We cast the problem of generating adversarial examples as an optimisation problem.
In particular, we propose a continuous \emph{inconsistency loss} that measures the \emph{degree} to which a set of sentences causes a model to violate a rule.

\begin{example}[Inconsistency Loss] \label{ex:inconsistency}
Consider the rule $\xrule{2}$ in \cref{tab:rules}, \ie $\rel{con}(\var_{1}, \var_{2}) \Rightarrow \rel{con}(\var_{2}, \var_{1})$.
Let $s_{1}, s_{2} \in \sspace$ be two sentences: this rule is violated if, according to the model, a sentence $s_{1}$ contradicts $s_{2}$, but $s_{2}$ does not contradict $s_{1}$.
However, if we just use the final decision made by the neural NLI model, we can simply check whether the rule is violated by two given sentences, without any information on the \emph{degree} of such a violation.
Intuitively, for the rule being \emph{maximally violated}, the conditional probability associated to $\rel{con}(s_{1}, s_{2})$ should be \emph{very high} ($\approx 1$), while the one associated to $\rel{con}(s_{2}, s_{1})$ should be \emph{very low} ($\approx 0$).
We can measure the extent to which the rule is violated -- which we refer to as \emph{inconsistency loss} $\iloss$ -- by checking whether the probability of the body of the rule is higher than the probability of its head:
\begin{equation*}
 \begin{aligned}
  \iloss(\sset = \{ & \var_{1} \mapsto s_{1}, \var_{2} \mapsto s_{2} \}) \\
  & = \left[ p_{\params}(\rel{con} \mid s_{1}, s_{2}) - p_{\params}(\rel{con} \mid s_{2}, s_{1}) \right]_{+}
 \end{aligned}
\end{equation*}
\noindent where $S$ is a \emph{substitution set} that maps the variables $X_{1}$ and $X_{2}$ in $\xrule{2}$ to the sentences $s_{1}$ and $s_{2}$, $[x]_{+} = \max(0, x)$, and $p_{\params}(\rel{con} \mid s_{i}, s_{j})$ is the (conditional) probability that $s_{i}$ contradicts $s_{j}$ according to the neural NLI model.
Note that, in accordance with the logic implication, the inconsistency loss reaches its global minimum when the probability of the body is close to zero -- \ie the \emph{premise} is false -- and when the probabilities of both the body and the head are close to one -- \ie the \emph{premise} and the \emph{conclusion} are both true.
\end{example}
We now generalise the intuition in Ex.~\ref{ex:inconsistency} to any FOL rule.
Let $r = (\rel{body} \Rightarrow \rel{head})$ denote an arbitrary FOL rule in the form described in \cref{eq:rule}, and let $\rel{vars}(r) = \{ X_{1}, \ldots, X_{n} \}$ denote the set of universally quantified variables in the rule $r$.
Furthermore, let $\sset = \{ X_{1} \mapsto s_{1}, \ldots, X_{n} \mapsto s_{n} \}$ denote a \emph{substitution set}, \ie a mapping from variables in $\rel{vars}(r)$ to sentences $s_{1}, \ldots, s_{n} \in \sspace$.
The inconsistency loss associated with the rule $r$ on the substitution set $\sset$ can be defined as:
\begin{equation} \label{eq:iloss}
 \begin{aligned}
  \iloss(\sset) = \left[ p(\sset ; \rel{body}) - p(\sset ; \rel{head}) \right]_{+}
 \end{aligned}
\end{equation}
\noindent where $p(\sset ; \rel{body})$ and $p(\sset ; \rel{head})$ denote the probability of body and head of the rule, after replacing the variables in $r$ with the corresponding sentences in $\sset$.
The motivation for the loss in \cref{eq:iloss} is that logic implications can be understood as ``whenever the body is true, the head has to be true as well''.
In terms of NLI models, this translates as ``the probability of the head should at least be as large as the probability of the body''.
For calculating the inconsistency loss in \cref{eq:iloss}, we need to specify how to calculate the probability of \rel{head} and \rel{body}.
The probability of a single ground atom is given by querying the neural NLI model, as in \cref{eq:nnli}.
The head contains a single atom, while the body can be a conjunction of multiple atoms.
Similarly to \citet{DBLP:conf/uai/MinerviniDRR17}, we use the G{\"o}del t-norm, a continuous generalisation of the conjunction operator in logic~\citep{Gupta:1991:TTN:107687.107690}, for computing the probability of the body of a clause:
\begin{equation*}
 p_{\params}(a_{1} \land a_{2}) = \min\{ p_{\params}(a_{1}), p_{\params}(a_{2}) \}
\end{equation*}
\noindent where $a_{1}$ and $a_{2}$ are two clause atoms.
In this work, we cast the problem of generating adversarial examples as an optimisation problem: we search for the substitution set $\sset = \{ X_{1} \mapsto s_{1}, \ldots, X_{n} \mapsto s_{n} \}$ that maximises the inconsistency loss in \cref{eq:iloss}, thus (maximally) violating the available background knowledge.
\subsection{Constraining via Language Modelling} \label{ssec:language}
Maximising the inconsistency loss in \cref{eq:iloss} may not be sufficient for generating meaningful adversarial examples: they can lead neural NLI models to violate available background knowledge, but they may not be well-formed and meaningful.
For such a reason, in addition to maximising the inconsistency loss, we also constrain the \emph{perplexity} of generated sentences by using a neural language model~\citep{DBLP:conf/nips/BengioDV00}.
In this work, we use a LSTM~\citep{DBLP:journals/neco/HochreiterS97} neural language model $p_{\mathcal{L}}(w_{1}, \ldots, w_{t})$ for generating low-perplexity adversarial examples.
\subsection{Searching in a Discrete Space} \label{ssec:discrete}
As mentioned earlier in this section, we cast the problem of automatically generating adversarial examples -- \ie examples that cause NLI models to violate available background knowledge -- as an optimisation problem.
Specifically, we look for substitutions sets $\sset = \{ X_{1} \mapsto s_{1}, \ldots, X_{n} \mapsto s_{n} \}$ that jointly:
\begin{inparaenum}[\itshape a\upshape)]
 \item maximise the \emph{inconsistency loss} described in \cref{eq:iloss}, and
 \item are composed by sentences with a low perplexity, as defined by the neural language model in \cref{ssec:language}.
\end{inparaenum}
The search objective can be formalised by the following optimisation problem:
\begin{equation} \label{eq:dsearch}
 \begin{aligned}
  & \underset{\sset}{\text{maximise}}
  & & \iloss(\sset) \\
  & \text{subject to}
  & & \log \langm(\sset) \leq \tau \\
 \end{aligned}
\end{equation}
\noindent where $\log \langm(\sset)$ denotes the log-probability of the sentences in the substitution set $\sset$, and $\tau$ is a threshold on the perplexity of generated sentences.
For generating low-perplexity adversarial examples, we take inspiration from \citet{DBLP:journals/corr/abs-1709-08878} and generate the sentences by editing prototypes extracted from a corpus.
Specifically, for searching substitution sets whose sentences jointly have a high probability and are highly adversarial, as measured the inconsistency loss in \cref{eq:iloss}, we use the following procedure: 
\begin{inparaenum}[\itshape a)\upshape]
 \item we first sample sentences close to the data manifold (\ie with a low perplexity), by either sampling from the training set or from the language model;
 \item we then make small variations to the sentences -- analogous to adversarial images, which consist in small perturbations of training examples -- so to optimise the objective in \cref{eq:dsearch}.
\end{inparaenum}
When editing prototypes, we consider the following perturbations:
\begin{inparaenum}[\itshape a)\upshape]
 \item change one word in one of the input sentences;
 \item remove one parse sub-tree from one of the input sentences;
 \item insert one parse sub-tree from one sentence in the corpus in the parse tree of one of the input sentences.
\end{inparaenum}
Note that the generation process can easily lead to ungrammatical or implausible sentences; however, these will be likely to have a high perplexity according to the language model (\cref{ssec:language}), and thus they will be ruled out by the search algorithm.
\section{Adversarial Regularisation} \label{sec:regularisation}
We now show one can use the adversarial examples to regularise the training process.
We propose training NLI models by jointly:
\begin{inparaenum}[\itshape a)\upshape]
 \item minimising the data loss (\cref{eq:loss}), and
 \item minimising the inconsistency loss (\cref{eq:iloss}) on a set of generated adversarial examples (substitution sets).
\end{inparaenum}
More formally, for training, we jointly minimise the cross-entropy loss defined on the data $\dloss(\params)$ and the inconsistency loss on a set of generated adversarial examples $\max_{\sset} \iloss(\sset ; \params)$, resulting in the following optimisation problem:
\begin{equation} \label{eq:jloss}
\begin{aligned}
 & \underset{\params}{\text{minimise}}
 & & \dloss(\data, \params) + \lambda \max_{\sset} \iloss(\sset ; \params) \\
  & \text{subject to}
  & & \log \langm(\sset) \leq \tau
\end{aligned}
\end{equation} 
\noindent where $\lambda \in \Real_{+}$ is a hyperparameter specifying the trade-off between the data loss $\dloss$ (\cref{eq:loss}), and the inconsistency loss $\iloss$ (\cref{eq:iloss}), measured on the generated substitution set $S$.
In \cref{eq:jloss}, the regularisation term $\max_{\sset} \iloss(\sset ; \params)$ has the task of \emph{generating} the adversarial substitution sets by maximising the inconsistency loss.
Furthermore, the constraint $\log \langm(\sset) \leq \tau$ ensures that the perplexity of generated sentences is lower than a threshold $\tau$.
For this work, we used the $\max$ aggregation function.
However, other functions can be used as well, such as the sum or mean of multiple inconsistency losses.
\begin{algorithm}[t]
 \caption{Solving the optimisation problem in \cref{eq:jloss} via Mini-Batch Gradient Descent} \label{alg:opt}
 \begin{algorithmic}[1]
 \REQUIRE{Dataset $\data$, weight $\lambda \in \Real_{+}$}
 \REQUIRE{No. of epochs $\tau \in \Natural_{+}$}
 \REQUIRE{No. of adv. substitution sets $n_{a} \in \Natural_{+}$}
  \STATE{\COMMENT{Initialise the model parameters $\hat{\params}$}}
  \STATE{$\hat{\params} \leftarrow \text{initialise}()$}
  \FOR{$i \in \{ 1, \ldots, \tau \}$}
   \FOR{$\data_{j} \in \text{batches}(\data)$}
    \STATE{\COMMENT{Generate the adv. substitution sets $S_{i}$}}
    \STATE{$\{ \sset_{1}, \ldots, \sset_{n_{a}} \} \leftarrow \text{generate}(\data_{j})$} \label{alg:opt:gen}
    \STATE{\COMMENT{Compute the gradient of \cref{eq:jloss}}}
    \STATE{$\mathcal{L} \leftarrow \dloss(\data_{j}, \hat{\params}) + \lambda \sum_{k = 1}^{n_{a}} \iloss(S_{k}; \hat{\params})$}
    \STATE{$g \leftarrow \nabla_{\params} \mathcal{L}$} \label{alg:opt:grad1}
    \STATE{\COMMENT{Update the model parameters}}
    \STATE{$\hat{\params} \leftarrow \hat{\params} - \eta g$} \label{alg:opt:update}
   \ENDFOR
  \ENDFOR
  \STATE{\textbf{return} $\hat{\params}$}
  \end{algorithmic}
\end{algorithm}
For minimising the regularised loss in \cref{eq:jloss}, we alternate between two optimisation processes -- generating the adversarial examples (\cref{eq:dsearch}) and minimising the regularised loss (\cref{eq:jloss}).
The algorithm is outlined in \cref{alg:opt}.
At each iteration, after generating a set of adversarial examples $\sset$, it computes the gradient of the regularised loss in \cref{eq:jloss}, and updates the model parameters via a gradient descent step.
On line \ref{alg:opt:gen}, the algorithm generates a set of adversarial examples, each in the form of a substitution set $\sset$.
On line \ref{alg:opt:grad1}, the algorithm computes the gradient of the adversarially regularised loss -- a weighted combination of the data loss in \cref{eq:loss} and the inconsistency loss in \cref{eq:iloss}.
The model parameters are finally updated on line \ref{alg:opt:update} via a gradient descent step.
\section{Creating Adversarial NLI Datasets} \label{sec:dataset}
\begin{table}[t]
 \centering
 \resizebox{\columnwidth}{!}{%
 \begin{tabular}{rl}
 \toprule
  {\bf Premise} & A man in a suit walks through a train station. \\
  {\bf Hypothesis} & Two boys ride skateboard. \\
  {\bf Type} & {\bf Contradiction} \\
 \midrule
  {\bf Premise} & Two boys ride skateboard. \\
  {\bf Hypothesis} & A man in a suit walks through a train station. \\
  {\bf Type} & {\bf Contradiction} \\
 \midrule
  {\bf Premise} & Two people are surfing in the ocean. \\
  {\bf Hypothesis} & There are people outside. \\
  {\bf Type} & {\bf Entailment} \\
 \midrule
  {\bf Premise} & There are people outside. \\
  {\bf Hypothesis} & Two people are surfing in the ocean. \\
  {\bf Type} & {\bf Neutral} \\
 \bottomrule
 \end{tabular}
 }
\caption{Sample sentences from an Adversarial NLI Dataset generated using the DAM model, by maximising the inconsistency loss $\iloss$.} \label{tab:adversarial}
\end{table}
We crafted a series of datasets for assessing the robustness of the proposed regularisation method to adversarial examples.
Starting from the SNLI test set, we proceeded as follows.
We selected the $k$ instances in the SNLI test set that maximise the inconsistency loss in \cref{eq:iloss} with respect to the rules in $\xrule{1}$, $\xrule{2}$, $\xrule{3}$, and $\xrule{4}$ in \cref{tab:rules}.
We refer to the generated datasets as $\aset{m}{k}$, where $m$ identifies the model used for selecting the sentence pairs, and $k$ denotes number of examples in the dataset.
For generating each of the $\aset{m}{k}$ datasets, we proceeded as follows.
Let $\mathcal{D} = \{ (x_{1}, y_{i}), \ldots, (x_{n}, y_{n}) \}$ be a NLI dataset (such as SNLI), where each instance $x_{i} = (p_{i}, h_{i})$ is a premise-hypothesis sentence pair, and $y_{i}$ denotes the relationship holding between $p_{i}$ and $h_{i}$.
For each instance $x_{i} = (p_{i}, h_{i})$, we consider two substitution sets: $\sset_{i} = \{ X_{1} \mapsto p_{i}, X_{2} \mapsto h_{i} \}$ and $\sset_{i}' = \{ X_{1} \mapsto h_{i}, X_{2} \mapsto p_{i} \}$, each corresponding to a mapping from variables to sentences.
We compute the \emph{inconsistency score} associated to each instance $x_{i}$ in the dataset $\mathcal{D}$ as $\iloss(\sset_{i}) + \iloss(\sset_{i}')$.
Note that the inconsistency score only depends on the premise $p_{i}$ and hypothesis $h_{i}$ in each instance $x_{i}$, and it does not depend on its label $y_{i}$.
After computing the inconsistency scores for all sentence pairs in $\mathcal{D}$ using a model $m$, we select the $k$ instances with the highest inconsistency score, we create two instances $x_{i} = (p_{i}, h_{i})$ and $\hat{x_{i}} = (h_{i}, p_{i})$, and add both $(x_{i}, y_{i})$ and $(\hat{x}_{i}, \hat{y}_{i})$ to the dataset $\aset{m}{k}$.
Note that, while $y_{i}$ is already known from the dataset $\mathcal{D}$, $\hat{y}_{i}$ is unknown.
For this reason, we find $\hat{y}_{i}$ by manual annotation.
\section{Related Work} \label{sec:related}
Adversarial examples are receiving a considerable attention in NLP; their usage, however, is considerably limited by the fact that semantically invariant input perturbations in NLP are difficult to identify~\citep{DBLP:journals/corr/BuckBCGHGW17}.
\citet{DBLP:conf/emnlp/JiaL17} analyse the robustness of extractive question answering models on examples obtained by adding adversarially generated distracting text to SQuAD~\citep{DBLP:conf/emnlp/RajpurkarZLL16} dataset instances.
\citet{DBLP:journals/corr/abs-1711-02173} also notice that character-level Machine Translation are overly sensitive to random character manipulations, such as typos.
\citet{DBLP:conf/cvpr/HosseiniXP17} show that simple character-level modifications can drastically change the toxicity score of a text.
\citet{DBLP:journals/corr/abs-1804-06059} proposes using paraphrasing for generating adversarial examples.
Our model is fundamentally different in two ways:
\begin{inparaenum}[\itshape a)\upshape]
 \item it does not need labelled data for generating adversarial examples -- the inconsistency loss can be maximised by just making an NLI model produce inconsistent results, and
 \item it incorporates adversarial examples during the training process, with the aim of training more robust NLI models.
\end{inparaenum}
\begin{table}[t]
 \centering
 \resizebox{\columnwidth}{!}{%
 \begin{tabular}{crC{1.2cm}C{1.2cm}C{1.2cm}C{1.2cm}}
  \toprule
\multicolumn{2}{c}{\bf Model} & \multicolumn{2}{c}{\bf Original}  & \multicolumn{2}{c}{\bf Regularised} \\
& & Valid. & Test & Valid. & Test \\
\midrule
\multirow{3}{*}{\begin{turn}{90}{\bf{\small MultiNLI}}\end{turn}} & \cBiLSTM & 61.52 & 63.95 & {\bf 66.98} & {\bf 66.68} \\
& \DAM & 72.78 & 73.28 & {\bf 73.57} & {\bf 73.51} \\
& \ESIM & 73.66 & 75.22 & {\bf 75.72} & {\bf 75.80} \\

\midrule

\multirow{3}{*}{\begin{turn}{90}{\bf{\small SNLI}}\end{turn}} & \cBiLSTM & 81.41 & 80.99 & {\bf 82.27} & {\bf 81.12} \\
& \DAM & 86.96 & 86.29 & {\bf 87.08} & {\bf 86.43} \\
& \ESIM & 87.83 & 87.25 & {\bf 87.98} & {\bf 87.55} \\
\bottomrule
 \end{tabular}
 }%
\caption{Accuracy on the SNLI and MultiNLI datasets with different neural NLI models \emph{before} (left) and \emph{after} (right) adversarial regularisation.} \label{tab:models}
\end{table}
Adversarial examples are also used for assessing the robustness of computer vision models~\citep{42503,DBLP:journals/corr/GoodfellowSS14,DBLP:conf/cvpr/NguyenYC15}, where they are created by adding a small amount of noise to the inputs that does not change the semantics of the images, but drastically changes the model predictions.
\section{Experiments} \label{sec:experiments}
\begin{table}[t]
    \centering
    \resizebox{\columnwidth}{!}{%
    \begin{tabular}{ccccc}

\toprule
{\bf Model} & {\bf Rule} & $\card{\mathbf{B}}$ & $\card{\mathbf{B} \land \lnot{\mathbf{H}}}$ & {\bf Violations} (\%) \\

\midrule

\multirow{4}{*}{\cBiLSTM{}} & $\xrule{1}$ & 1,098,734 & 261,064 & 23.76 \% \\
& $\xrule{2}$ & 174,902 & 80,748 & 46.17 \% \\
& $\xrule{3}$ & 197,697 & 24,294 & 12.29 \% \\
& $\xrule{4}$ & 176,768 & 33,435 & 18.91 \% \\

\midrule

\multirow{4}{*}{\DAM{}} & $\xrule{1}$ & 1,098,734 & 956 & 00.09 \% \\
& $\xrule{2}$ & 171,728 & 28,680 & 16.70 \% \\
& $\xrule{3}$ & 196,042 & 11,599 & 05.92 \% \\
& $\xrule{4}$ & 181,597 & 29,635 & 16.32 \% \\

\midrule

\multirow{4}{*}{\ESIM{}} & $\xrule{1}$ & 1,098,734 & 10,985 & 01.00 \% \\
& $\xrule{2}$ & 177,950 & 17,518 & 09.84 \% \\
& $\xrule{3}$ & 200,852 & 6,482 & 03.23 \% \\
& $\xrule{4}$ & 170,565 & 17,190 & 10.08 \% \\
     
\bottomrule

    \end{tabular}
    }%
\caption{Violations (\%) of rules $\xrule{1}, \xrule{2}, \xrule{3}, \xrule{4}$ from \cref{tab:rules} on the SNLI training set, yield by \cBiLSTM{}, \DAM{}, and \ESIM{}.}
    \label{tab:violations}
\end{table}
We trained \DAM{}, \ESIM{} and \cBiLSTM{} on the SNLI corpus using the hyperparameters provided in the respective papers.
The results provided by such models on the SNLI and MultiNLI validation and tests sets are provided in \cref{tab:models}.
In the case of MultiNLI, the validation set was obtained by removing 10,000 instances from the training set (originally composed by 392,702 instances), and the test set consists in the \emph{matched} validation set.
\begin{table*}[t]
 \centering
 \resizebox{\textwidth}{!}{
 \centering
    
\begin{tabular}{R{3cm}C{1.5cm}C{1.5cm}C{1.5cm}C{1.5cm}C{1.5cm}C{1.5cm}C{1.5cm}C{1.5cm}C{1.5cm}}
\toprule
\diagbox{\bf Model}{\bf Dataset} & $\aset{DAM}{100}$ & $\aset{DAM}{500}$ & $\aset{DAM}{1000}$ & $\aset{ESIM}{100}$ & $\aset{ESIM}{500}$ & $\aset{ESIM}{1000}$ & $\aset{cBiLSTM}{100}$ & $\aset{cBiLSTM}{500}$ & $\aset{cBiLSTM}{1000}$ \\

\cmidrule(lr){1-1}
\cmidrule(lr){2-4}
\cmidrule(lr){5-7}
\cmidrule(lr){8-10}

\DAM$^{\mathcal{AR}}$ & {\bf 83.33} & {\bf 79.15} & {\bf 79.37} & {\bf 71.35} & {\bf 72.19} & {\bf 70.05} & {\bf 93.00} & {\bf 88.99} & {\bf 86.00} \\ 
\DAM & 47.40 & 47.93 & 51.66 & 55.73 & 60.94 & 60.88 & 81.50 & 77.37 & 75.28 \\

    \cmidrule(lr){1-1}
    \cmidrule(lr){2-4}
    \cmidrule(lr){5-7}
    \cmidrule(lr){8-10}

    \ESIM$^{\mathcal{AR}}$ & {\bf 89.06} & {\bf 86.00} & {\bf 85.08} & {\bf 78.12} & {\bf 76.04} & {\bf 73.32} & {\bf 96.50} & {\bf 91.92} & {\bf 88.52} \\ 
\ESIM & 72.40 & 74.59 & 76.92 & 52.08 & 58.65 & 60.78 & 87.00 & 84.34 & 82.05 \\

\cmidrule(lr){1-1}
\cmidrule(lr){2-4}
\cmidrule(lr){5-7}
\cmidrule(lr){8-10}

\cBiLSTM$^{\mathcal{AR}}$ & {\bf 85.42} & {\bf 80.39} & {\bf 78.74} & {\bf 73.96} & {\bf 70.52} & {\bf 65.39} & {\bf 92.50} & {\bf 88.38} & {\bf 83.62} \\ 
\cBiLSTM & 56.25 & 59.96 & 61.75 & 47.92 & 53.23 & 53.73 & 51.50 & 52.83 & 53.24 \\ 

\bottomrule
\end{tabular}

 }
 \caption{Accuracy of unregularised and regularised neural NLI models \DAM{}, \cBiLSTM{}, and \ESIM{}, and their adversarially regularised versions \DAM$^{\mathcal{AR}}$, \cBiLSTM$^{\mathcal{AR}}$, and \ESIM$^{\mathcal{AR}}$, on the datasets $\aset{m}{k}$.} \label{tab:accuracy}
\end{table*}
\paragraph{\bf Background Knowledge Violations.}
As a first experiment, we count the how likely our model is to violate rules $\xrule{1}, \xrule{2}, \xrule{3}, \xrule{4}$ in \cref{tab:rules}.
In \cref{tab:violations} we report the number sentence pairs in the SNLI training set where \DAM{}, \ESIM{} and \cBiLSTM{} violate $\xrule{1}, \xrule{2}, \xrule{3}, \xrule{4}$.
In the $\card{\mathbf{B}}$ column we report the number of times the body of the rule holds, according to the model.
In the $\card{\mathbf{B} \land \lnot{\mathbf{H}}}$ column we report the number of times where the body of the rule holds, but the head does not -- which is clearly a violation of available rules.
We can see that, in the case of rule $\xrule{1}$ (reflexivity of entailment), \DAM{} and \ESIM{} make a relatively low number of violations -- namely 0.09 and 1.00 \%, respectively.
However, in the case of \cBiLSTM{}, we can see that, each sentence $s \in \sspace$ in the SNLI training set, with a 23.76 \% chance, $s$ does not entail itself -- which violates our background knowledge.
With respect to $\xrule{2}$ (symmetry of contradiction), we see that none of the models is completely consistent with the available background knowledge.
Given a sentence pair $s_{1}, s_{2} \in \sspace$ from the SNLI training set, if -- according to the model -- $s_{1}$ contradicts $s_{2}$, a significant number of times (between 9.84\% and 46.17\%) the same model also infers that $s_{2}$ \emph{does not} contradict $s_{1}$.
This phenomenon happens 16.70 \% of times with \DAM{}, 9.84 \% of times with \ESIM{}, and 46.17 \% with \cBiLSTM{}: this indicates that all considered models are prone to violating $\xrule{2}$ in their predictions, with \ESIM{} being the more robust.
In \cref{sec:a:adversarial} we report several examples of such violations in the SNLI training set.
We select those that maximise the inconsistency loss described in \cref{eq:iloss}, violating rules $\xrule{2}$ and $\xrule{3}$.
We can notice that the presence of inconsistencies is often correlated with the length of the sentences.
The model tends to detect entailment relationships between longer (\ie, possibly more specific) and shorter (\ie, possibly more general) sentences.

\subsection{Generation of Adversarial Examples}
In the following, we analyse the automatic generation of sets of adversarial examples that make the model violate the existing background knowledge.
We search in the space of sentences by applying perturbations to sampled sentence pairs, using a language model for guiding the search process.
The generation procedure is described in \cref{sec:generating}.
\begin{table*}[t]
    \centering
    \resizebox{\textwidth}{!}{
\begin{tabular}{lrlcc}
\toprule
& & {\bf Adversarial Example} & {\bf Prediction} & {\bf Inconsistency} \\

\midrule

\multirow{2}{*}{1} & $s_{1}$ & A man in uniform is pushing a medical bed. & \cfbox{red}{ $s_{1} \textcolor{green}{\xrightarrow{0.72}} s_{2}$ } & \multirow{2}{*}{$.01 \rightsquigarrow .92$} \\
& $s_{2}$ & a man is \sout{pushing} \textcolor{red}{carrying} something. & $s_{2} \textcolor{red}{\xrightarrow{0.93}} s_{1}$ & \\

\midrule

\multirow{2}{*}{1} & $s_{1}$ & A dog swims in the water & $s_{1} \textcolor{green}{\xrightarrow{0.78}} s_{2}$ & \multirow{2}{*}{$.00 \rightsquigarrow .99$} \\
& $s_{2}$ & A dog is \sout{swimming} outside. & \cfbox{red}{ $s_{2} \textcolor{red}{\xrightarrow{0.99}} s_{1}$ } & \\

\midrule

\multirow{2}{*}{2} & $s_{1}$ & A young man is sledding down a snow covered hill on a green sled. & $s_{1} \textcolor{blue}{\xrightarrow{0.98}} s_{2}$ & \multirow{2}{*}{$.00 \rightsquigarrow .97$} \\
& $s_{1}$ & A man is \sout{sledding} down to meet his daughter. & \cfbox{red}{ $s_{2} \textcolor{red}{\xrightarrow{1.00}} s_{1}$ } & \\

\midrule



\multirow{2}{*}{3} & $s_{1}$ & \textcolor{red}{A woman sleeps on the ground.} A boy and girl play in a pool. & $s_{1} \textcolor{blue}{\xrightarrow{0.94}} s_{2}$ & \multirow{2}{*}{$.00 \rightsquigarrow .82$} \\
& $s_{2}$ & Two kids are happily playing in a swimming pool. & \cfbox{red}{ $s_{2} \textcolor{red}{\xrightarrow{0.85}} s_{1}$ } & \\


\midrule

\multirow{2}{*}{4} & $s_{1}$ & The school is having a special event in order to show the american culture on how other cultures are dealt with in parties. & $s_{1} \textcolor{red}{\xrightarrow{0.96}} s_{2}$ & \multirow{2}{*}{$.01 \rightsquigarrow .63$} \\
& $s_{2}$ & A \sout{school} \textcolor{red}{dog} is hosting an event. & \cfbox{red}{ $s_{2} \textcolor{blue}{\xrightarrow{0.66}} s_{1}$ } & \\

\midrule

& $s_{1}$ & A boy is drinking out of a water fountain shaped like a woman. & $s_{1} \textcolor{green}{\xrightarrow{0.96}} s_{2}$ & \\
5 & $s_{2}$ & A male is getting a drink of water. & \cfbox{red}{ $s_{2} \textcolor{green}{\xrightarrow{0.93}} s_{3}$ } & $.00 \rightsquigarrow .94$ \\
& $s_{3}$ & A \sout{male} \textcolor{red}{man} is getting a drink of water. & $s_{1} \textcolor{red}{\xrightarrow{0.97}} s_{3}$ & \\

\bottomrule
\end{tabular}
    }
\caption{
    Inconsistent results produced by \DAM{} on automatically generated adversarial examples.
    The notation \sout{segment one} \textcolor{red}{segment two} denotes that the corruption process removes ``segment one'' and introduced ``segment two'' in the sentence, and $s_{1} \textcolor{red}{\xrightarrow{p}} s_{2}$ indicates that \DAM{} classifies the relation between $s_{1}$ and $s_{2}$ as \emph{contradiction}, with probability $p$.
    We use different colours for representing the \textcolor{red}{contradiction}, \textcolor{green}{entailment} and \textcolor{blue}{neutral} classes.
    Examples 1, 2, 3, and 4 violate the rule $\xrule{2}$, while example 5 violates the rule $\xrule{5}$. $.00 \rightsquigarrow .99$ indicates that the corruption process increases the inconsistency loss from .00 to .99, and the \cfbox{red}{red boxes} are used for indicating mistakes made by the model on the adversarial examples.
} \label{tab:corruptions_short}
\end{table*}
\begin{figure}[t]
 \centering
 \includegraphics[width=\columnwidth]{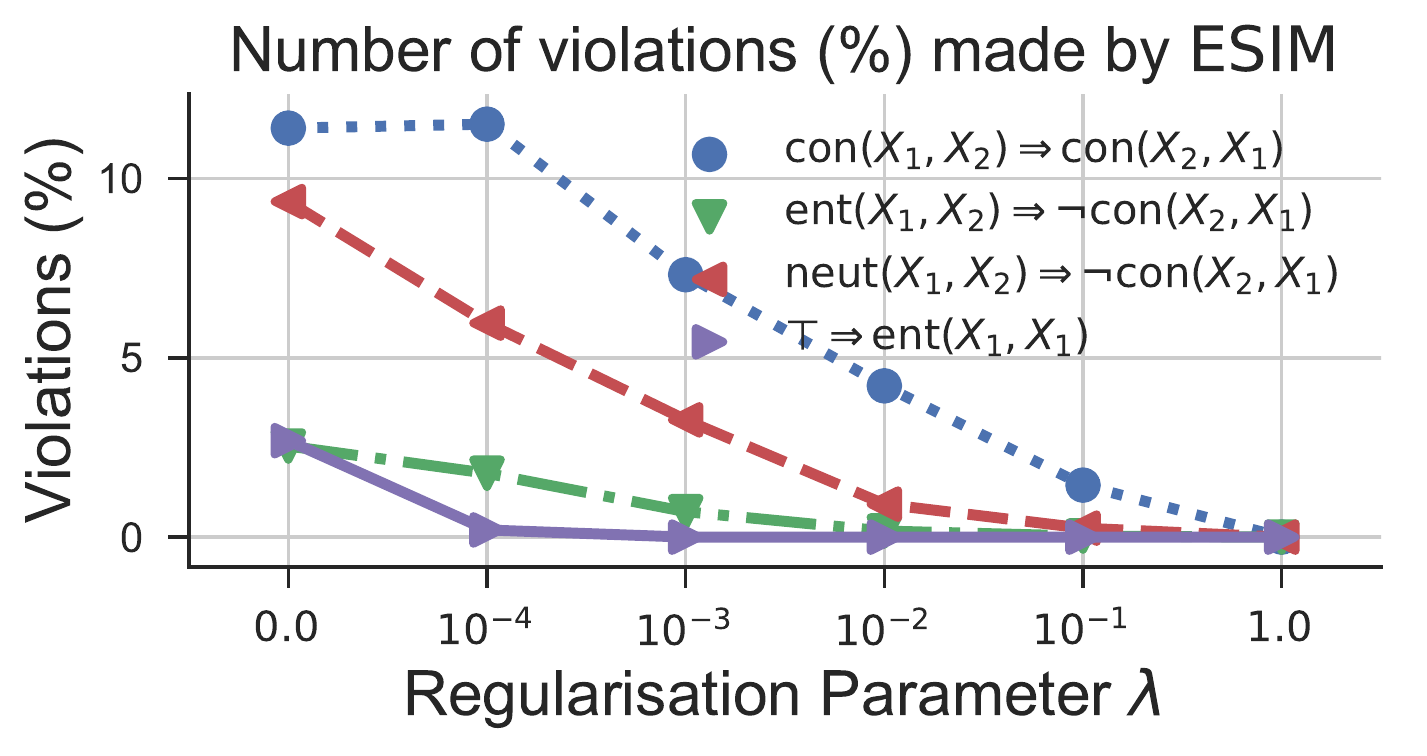}
 \caption{Number of violations (\%) to rules in \cref{tab:rules} made by \ESIM{} on the SNLI test set.} \label{fig:violations}
\end{figure}
The procedure was especially effective in generating adversarial examples -- a sample is shown in \cref{tab:corruptions_short}.
We can notice that, even though \DAM{} and \ESIM{} achieve results close to human level performance on SNLI, they are likely to fail when faced with linguistic phenomena such as negation, hyponymy, and antonymy.
\citet{DBLP:journals/corr/abs-1803-02324} recently showed that NLI datasets tend to suffer from annotation artefacts and limited linguistic variations: this allows NLI models to achieve nearly-human performance by capturing repetitive patterns and idiosyncrasies in a dataset, without being able of effectively capturing textual entailment.
This is visible, for instance, in example 5 of \cref{tab:corruptions_short}, where the model fails to capture the hyponymy relation between ``male'' and ``man'', incorrectly predicting an entailment in place of a neutral relationship.
Furthermore, it is clear that models lack commonsense knowledge, such as the relation between ``pushing'' and ``carrying'' (example 1), and being outside and swimming (example 2).
Generating such adversarial examples provides us with useful insights on the inner workings of neural NLI models, that can be leveraged for improving the robustness of state-of-the-art models.
\subsection{Adversarial Regularisation}
We evaluated whether our approach for integrating logical background knowledge via adversarial training (\cref{sec:regularisation}) is effective at reducing the number of background knowledge violations, without reducing the predictive accuracy of the model.
We started with pre-trained \DAM{}, \ESIM{}, and \cBiLSTM{} models, trained using the hyperparameters published in their respective papers.
After training, each model was then fine-tuned for 10 epochs, by minimising the adversarially regularised loss function introduced in \cref{eq:jloss}.
\cref{tab:models} shows results on the SNLI and MultiNLI development and test set, while \cref{fig:violations} shows the number of violations for different values of $\lambda$, where regularised models are much more likely to make predictions that are consistent with the available background knowledge.
We can see that, despite the drastic reduction of background knowledge violations, the improvement may not be significant, supporting the idea that models achieving close-to-human performance on SNLI and MultiNLI may be capturing annotation artefacts and idiosyncrasies in such datasets~\citep{DBLP:journals/corr/abs-1803-02324}.
\paragraph{Evaluation on Adversarial Datasets.}
We evaluated the proposed approach on 9 adversarial datasets $\aset{m}{k}$, with $k \in \{ 100, 500, 1000 \}$, generated following the procedure described in \cref{sec:dataset} -- results are summarised in \cref{tab:accuracy}.
We can see that the proposed adversarial training method significantly increases the accuracy on the adversarial test sets.
For instance, consider $\aset{DAM}{100}$: prior to regularising ($\lambda = 0$), \DAM{} achieves a very low accuracy on this dataset -- \ie $47.4 \%$.
By increasing the regularisation parameter $\lambda \in \{ 10^{-4}, 10^{-3}, 10^{-2}, 10^{-1} \}$, we noticed sensible accuracy increases, yielding relative accuracy improvements up to $75.8 \%$ in the case of \DAM{}, and $79.6 \%$ in the case of \cBiLSTM{}.
From \cref{tab:accuracy} we can notice that adversarial examples \emph{transfer} across different models: an unregularised model is likely to perform poorly also on adversarial datasets generated by using different models, with \ESIM{} being the more robust model to adversarially generated examples.
Furthermore, we can see that regularised models are generally more robust to adversarial examples, even when those were generated using different model architectures.
For instance we can see that, while \cBiLSTM{} is vulnerable also to adversarial examples generated using \DAM{} and \ESIM{}, its adversarially regularised version \cBiLSTM$^{\mathcal{AR}}$ is generally more robust to any sort of adversarial examples.
\section{Conclusions} \label{sec:conclusions}
In this paper, we investigated the problem of automatically generating adversarial examples that violate a set of given First-Order Logic constraints in NLI.
We reduced the problem of identifying such adversarial examples to an optimisation problem, by maximising a continuous relaxation of the violation of such constraints, and by using a language model for generating linguistically-plausible examples.
Furthermore, we proposed a method for adversarially regularising neural NLI models for incorporating background knowledge.
Our results showed that the proposed method consistently yields significant increases to the predictive accuracy on adversarially-crafted datasets -- up to a 79.6\% relative improvement -- while drastically reducing the number of background knowledge violations.
Furthermore, we showed that adversarial examples transfer across model architectures, and the proposed adversarial training procedure produces generally more robust models.
The source code and data for reproducing our results is available online, at \href{https://github.com/uclmr/adversarial-nli/}{https://github.com/uclmr/adversarial-nli/}.
\subsection*{Acknowledgements}
We are immensely grateful to Jeff Mitchell, Johannes Welbl, Sameer Singh, and the whole UCL Machine Reading group for all useful discussions, inputs, and ideas.
This work has been supported by an Allen Distinguished Investigator Award.
\bibliography{bibliography}

\begin{thebibliography}{33}
\expandafter\ifx\csname natexlab\endcsname\relax\def\natexlab#1{#1}\fi

\bibitem[{Belinkov and Bisk(2017)}]{DBLP:journals/corr/abs-1711-02173}
Yonatan Belinkov and Yonatan Bisk. 2017.
\newblock Synthetic and natural noise both break neural machine translation.
\newblock \emph{CoRR}, abs/1711.02173.

\bibitem[{Bengio et~al.(2000)Bengio, Ducharme, and
  Vincent}]{DBLP:conf/nips/BengioDV00}
Yoshua Bengio, R{\'{e}}jean Ducharme, and Pascal Vincent. 2000.
\newblock A neural probabilistic language model.
\newblock In \emph{Advances in Neural Information Processing Systems 13, Papers
  from Neural Information Processing Systems {(NIPS)} 2000}, pages 932--938.
  {MIT} Press.

\bibitem[{van Benthem(2008)}]{vanBenthem08NATLOG}
Johan van Benthem. 2008.
\newblock A brief history of natural logic.
\newblock In M.~Chakraborty, B.~L{\"o}we, M.~Nath~Mitra, and S.~Sarukki,
  editors, \emph{Logic, {N}avya-{N}yaya and Applications: Homage to {B}imal
  {M}atilal}. College Publications.

\bibitem[{Bos and Markert(2005)}]{DBLP:conf/naacl/BosM05}
Johan Bos and Katja Markert. 2005.
\newblock Recognising textual entailment with logical inference.
\newblock In \emph{{HLT/EMNLP} 2005, Human Language Technology Conference and
  Conference on Empirical Methods in Natural Language Processing, Proceedings
  of the Conference}, pages 628--635. The Association for Computational
  Linguistics.

\bibitem[{Bowman et~al.(2015)Bowman, Angeli, Potts, and
  Manning}]{DBLP:conf/emnlp/BowmanAPM15}
Samuel~R. Bowman, Gabor Angeli, Christopher Potts, and Christopher~D. Manning.
  2015.
\newblock A large annotated corpus for learning natural language inference.
\newblock In \emph{Proceedings of the 2015 Conference on Empirical Methods in
  Natural Language Processing, {EMNLP} 2015}, pages 632--642. The Association
  for Computational Linguistics.

\bibitem[{Buck et~al.(2017)Buck, Bulian, Ciaramita, Gesmundo, Houlsby,
  Gajewski, and Wang}]{DBLP:journals/corr/BuckBCGHGW17}
Christian Buck, Jannis Bulian, Massimiliano Ciaramita, Andrea Gesmundo, Neil
  Houlsby, Wojciech Gajewski, and Wei Wang. 2017.
\newblock Ask the right questions: Active question reformulation with
  reinforcement learning.
\newblock \emph{CoRR}, abs/1705.07830.

\bibitem[{Chen et~al.(2017)Chen, Zhu, Ling, Wei, Jiang, and
  Inkpen}]{DBLP:conf/acl/ChenZLWJI17}
Qian Chen, Xiaodan Zhu, Zhen{-}Hua Ling, Si~Wei, Hui Jiang, and Diana Inkpen.
  2017.
\newblock Enhanced {LSTM} for natural language inference.
\newblock In \emph{Proceedings of the 55th Annual Meeting of the Association
  for Computational Linguistics, {ACL} 2017}, pages 1657--1668. Association for
  Computational Linguistics.

\bibitem[{Condoravdi et~al.(2003)Condoravdi, Crouch, de~Paiva, Stolle, and
  Bobrow}]{Condoravdi-etAl:2003}
Cleo Condoravdi, Dick Crouch, Valeria de~Paiva, Reinhard Stolle, and Daniel~G.
  Bobrow. 2003.
\newblock Entailment, intensionality and text understanding.
\newblock In \emph{Proceedings of the HLT-NAACL 2003 Workshop on Text Meaning},
  pages 38--45.

\bibitem[{Dagan et~al.(2005)Dagan, Glickman, and
  Magnini}]{DBLP:conf/mlcw/DaganGM05}
Ido Dagan, Oren Glickman, and Bernardo Magnini. 2005.
\newblock The {PASCAL} recognising textual entailment challenge.
\newblock In \emph{Machine Learning Challenges, Evaluating Predictive
  Uncertainty, Visual Object Classification and Recognizing Textual Entailment,
  First {PASCAL} Machine Learning Challenges Workshop, {MLCW} 2005}, volume
  3944 of \emph{LNCS}, pages 177--190. Springer.

\bibitem[{Fyodorov et~al.(2000)Fyodorov, Winter, and
  Francez}]{Fyodorov-etal:2000}
Yaroslav Fyodorov, Yoad Winter, and Nissim Francez. 2000.
\newblock A natural logic inference system.
\newblock In \emph{Proceedings of the of the 2nd Workshop on Inference in
  Computational Semantics}.

\bibitem[{Goodfellow et~al.(2014)Goodfellow, Shlens, and
  Szegedy}]{DBLP:journals/corr/GoodfellowSS14}
Ian~J. Goodfellow, Jonathon Shlens, and Christian Szegedy. 2014.
\newblock Explaining and harnessing adversarial examples.
\newblock \emph{CoRR}, abs/1412.6572.

\bibitem[{Gupta and Qi(1991)}]{Gupta:1991:TTN:107687.107690}
M.~M. Gupta and J.~Qi. 1991.
\newblock Theory of t-norms and fuzzy inference methods.
\newblock \emph{Fuzzy Sets Syst.}, 40(3):431--450.

\bibitem[{Gururangan et~al.(2018)Gururangan, Swayamdipta, Levy, Schwartz,
  Bowman, and Smith}]{DBLP:journals/corr/abs-1803-02324}
Suchin Gururangan, Swabha Swayamdipta, Omer Levy, Roy Schwartz, Samuel~R.
  Bowman, and Noah~A. Smith. 2018.
\newblock Annotation artifacts in natural language inference data.
\newblock \emph{CoRR}, abs/1803.02324.

\bibitem[{Guu et~al.(2017)Guu, Hashimoto, Oren, and
  Liang}]{DBLP:journals/corr/abs-1709-08878}
Kelvin Guu, Tatsunori~B. Hashimoto, Yonatan Oren, and Percy Liang. 2017.
\newblock Generating sentences by editing prototypes.
\newblock \emph{CoRR}, abs/1709.08878.

\bibitem[{Hastie et~al.(2001)Hastie, Tibshirani, and
  Friedman}]{hastie01statisticallearning}
Trevor Hastie, Robert Tibshirani, and Jerome Friedman. 2001.
\newblock \emph{The Elements of Statistical Learning}.
\newblock Springer Series in Statistics. Springer New York Inc.

\bibitem[{Hochreiter and Schmidhuber(1997)}]{DBLP:journals/neco/HochreiterS97}
Sepp Hochreiter and J{\"{u}}rgen Schmidhuber. 1997.
\newblock Long short-term memory.
\newblock \emph{Neural Computation}, 9(8):1735--1780.

\bibitem[{Hosseini et~al.(2017)Hosseini, Xiao, and
  Poovendran}]{DBLP:conf/cvpr/HosseiniXP17}
Hossein Hosseini, Baicen Xiao, and Radha Poovendran. 2017.
\newblock Deceiving google's cloud video intelligence {API} built for
  summarizing videos.
\newblock In \emph{2017 {IEEE} Conference on Computer Vision and Pattern
  Recognition Workshops, {CVPR} Workshops}, pages 1305--1309. {IEEE} Computer
  Society.

\bibitem[{Iyyer et~al.(2018)Iyyer, Wieting, Gimpel, and
  Zettlemoyer}]{DBLP:journals/corr/abs-1804-06059}
Mohit Iyyer, John Wieting, Kevin Gimpel, and Luke Zettlemoyer. 2018.
\newblock Adversarial example generation with syntactically controlled
  paraphrase networks.
\newblock \emph{CoRR}, abs/1804.06059.

\bibitem[{Jia and Liang(2017)}]{DBLP:conf/emnlp/JiaL17}
Robin Jia and Percy Liang. 2017.
\newblock Adversarial examples for evaluating reading comprehension systems.
\newblock In \emph{Proceedings of the 2017 Conference on Empirical Methods in
  Natural Language Processing, {EMNLP} 2017}, pages 2011--2021. Association for
  Computational Linguistics.

\bibitem[{Kannan and Vinyals(2017)}]{DBLP:journals/corr/KannanV17}
Anjuli Kannan and Oriol Vinyals. 2017.
\newblock Adversarial evaluation of dialogue models.
\newblock \emph{CoRR}, abs/1701.08198.

\bibitem[{Katz(1972)}]{katz1972semantic}
J.J. Katz. 1972.
\newblock \emph{Semantic theory}.
\newblock Studies in language. Harper \& Row.

\bibitem[{Levesque(2014)}]{DBLP:journals/ai/Levesque14}
Hector~J. Levesque. 2014.
\newblock On our best behaviour.
\newblock \emph{Artif. Intell.}, 212:27--35.

\bibitem[{MacCartney and Manning(2009)}]{maccartney2009extended}
Bill MacCartney and Christopher~D Manning. 2009.
\newblock An extended model of natural logic.
\newblock In \emph{Proceedings of the of the Eighth International Conference on
  Computational Semantics}, Tilburg, Netherlands.

\bibitem[{Minervini et~al.(2017)Minervini, Demeester, Rockt{\"{a}}schel, and
  Riedel}]{DBLP:conf/uai/MinerviniDRR17}
Pasquale Minervini, Thomas Demeester, Tim Rockt{\"{a}}schel, and Sebastian
  Riedel. 2017.
\newblock Adversarial sets for regularising neural link predictors.
\newblock In \emph{Proceedings of the Thirty-Third Conference on Uncertainty in
  Artificial Intelligence, {UAI} 2017}. {AUAI} Press.

\bibitem[{Nguyen et~al.(2015)Nguyen, Yosinski, and
  Clune}]{DBLP:conf/cvpr/NguyenYC15}
Anh~Mai Nguyen, Jason Yosinski, and Jeff Clune. 2015.
\newblock Deep neural networks are easily fooled: High confidence predictions
  for unrecognizable images.
\newblock In \emph{{IEEE} Conference on Computer Vision and Pattern
  Recognition, {CVPR} 2015}, pages 427--436. {IEEE} Computer Society.

\bibitem[{Paperno et~al.(2016)Paperno, Kruszewski, Lazaridou, Pham, Bernardi,
  Pezzelle, Baroni, Boleda, and
  Fern{\'{a}}ndez}]{DBLP:conf/acl/PapernoKLPBPBBF16}
Denis Paperno, Germ{\'{a}}n Kruszewski, Angeliki Lazaridou, Quan~Ngoc Pham,
  Raffaella Bernardi, Sandro Pezzelle, Marco Baroni, Gemma Boleda, and Raquel
  Fern{\'{a}}ndez. 2016.
\newblock The {LAMBADA} dataset: Word prediction requiring a broad discourse
  context.
\newblock In \emph{Proceedings of the 54th Annual Meeting of the Association
  for Computational Linguistics, {ACL} 2016}. The Association for Computer
  Linguistics.

\bibitem[{Parikh et~al.(2016)Parikh, T{\"{a}}ckstr{\"{o}}m, Das, and
  Uszkoreit}]{DBLP:conf/emnlp/ParikhT0U16}
Ankur~P. Parikh, Oscar T{\"{a}}ckstr{\"{o}}m, Dipanjan Das, and Jakob
  Uszkoreit. 2016.
\newblock A decomposable attention model for natural language inference.
\newblock In  \cite{DBLP:conf/emnlp/2016}, pages 2249--2255.

\bibitem[{Rajpurkar et~al.(2016)Rajpurkar, Zhang, Lopyrev, and
  Liang}]{DBLP:conf/emnlp/RajpurkarZLL16}
Pranav Rajpurkar, Jian Zhang, Konstantin Lopyrev, and Percy Liang. 2016.
\newblock Squad: 100, 000+ questions for machine comprehension of text.
\newblock In  \cite{DBLP:conf/emnlp/2016}, pages 2383--2392.

\bibitem[{Rimell and Clark(2009)}]{DBLP:journals/jbi/RimellC09}
Laura Rimell and Stephen Clark. 2009.
\newblock Porting a lexicalized-grammar parser to the biomedical domain.
\newblock \emph{Journal of Biomedical Informatics}, 42(5):852--865.

\bibitem[{Rockt{\"{a}}schel et~al.(2016)Rockt{\"{a}}schel, Grefenstette,
  Hermann, Kocisky, and Blunsom}]{rocktaschel2016reasoning}
Tim Rockt{\"{a}}schel, Edward Grefenstette, Karl~Moritz Hermann, Tomas Kocisky,
  and Phil Blunsom. 2016.
\newblock Reasoning about entailment with neural attention.
\newblock In \emph{International Conference on Learning Representations
  (ICLR)}.

\bibitem[{Su et~al.(2016)}]{DBLP:conf/emnlp/2016}
Jian Su et~al., editors. 2016.
\newblock \emph{Proceedings of the 2016 Conference on Empirical Methods in
  Natural Language Processing, {EMNLP} 2016}. The Association for Computational
  Linguistics.

\bibitem[{Szegedy et~al.(2014)Szegedy, Zaremba, Sutskever, Bruna, Erhan,
  Goodfellow, and Fergus}]{42503}
Christian Szegedy, Wojciech Zaremba, Ilya Sutskever, Joan Bruna, Dumitru Erhan,
  Ian Goodfellow, and Rob Fergus. 2014.
\newblock Intriguing properties of neural networks.
\newblock In \emph{International Conference on Learning Representations}.

\bibitem[{Williams et~al.(2017)Williams, Nangia, and
  Bowman}]{DBLP:journals/corr/WilliamsNB17}
Adina Williams, Nikita Nangia, and Samuel~R. Bowman. 2017.
\newblock A broad-coverage challenge corpus for sentence understanding through
  inference.
\newblock \emph{CoRR}, abs/1704.05426.

\end{thebibliography}
\bibliographystyle{acl_natbib_nourl}
\clearpage
\appendix
\section{Supplementary Material} \label{sec:a:supplementary}
\subsection{Accuracy on Adversarial Datasets} \label{sec:a:accuracy}
In the following, we report the accuracy of \DAM{} on several adversarial datasets $\aset{m}{k}$, with $k = 100$ and $m \in \{ \DAM,$ $\ESIM,$ $\cBiLSTM \}$.
\includegraphics[width=\columnwidth]{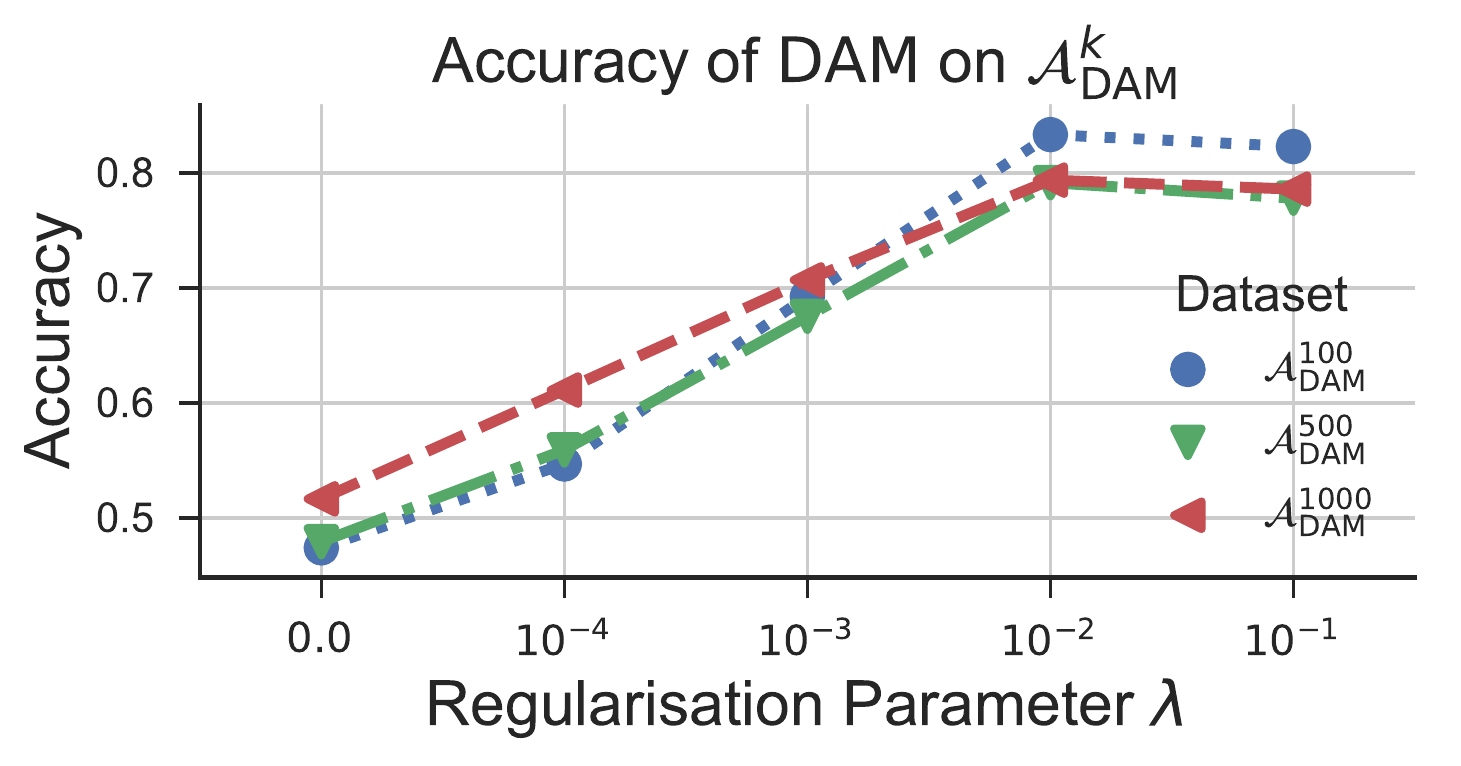}
\includegraphics[width=\columnwidth]{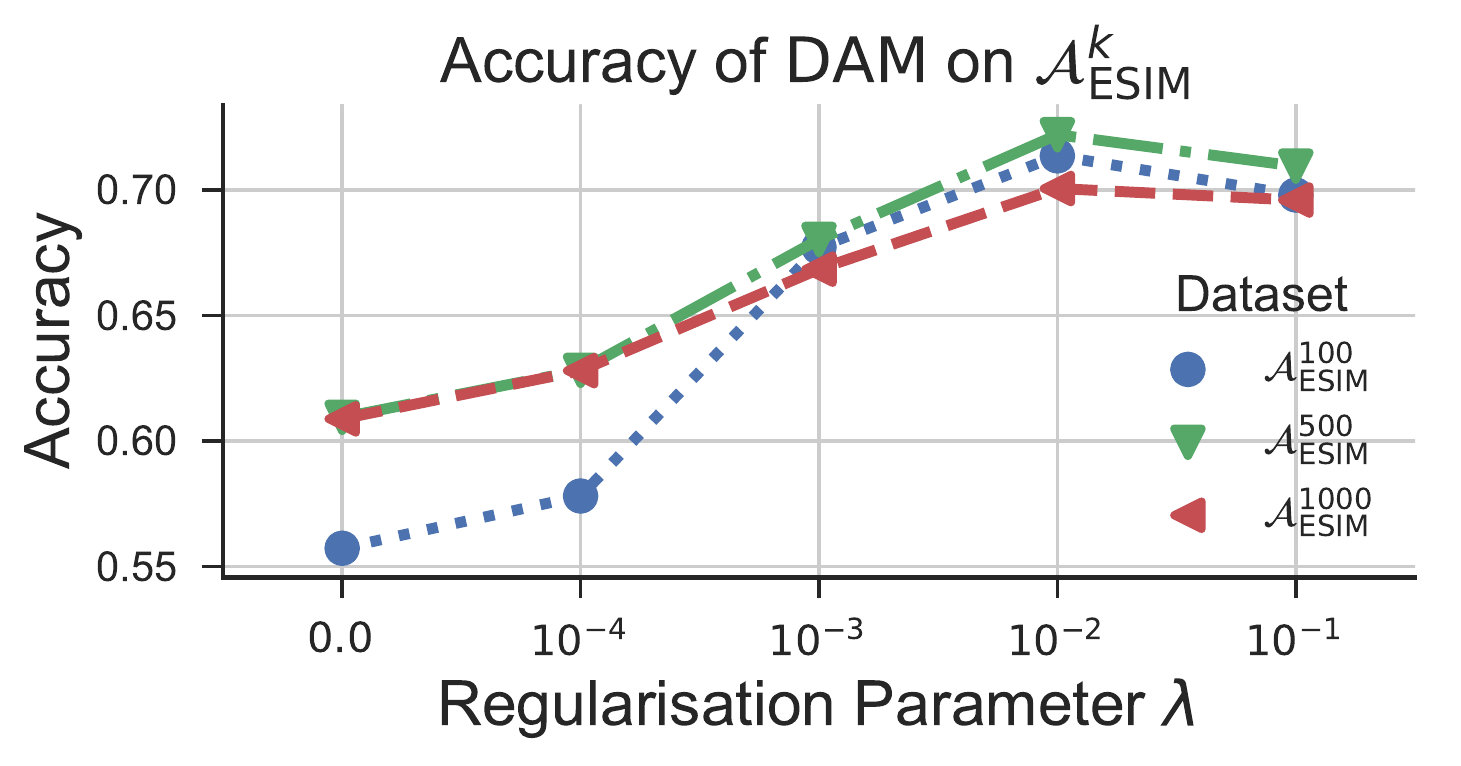}
\includegraphics[width=\columnwidth]{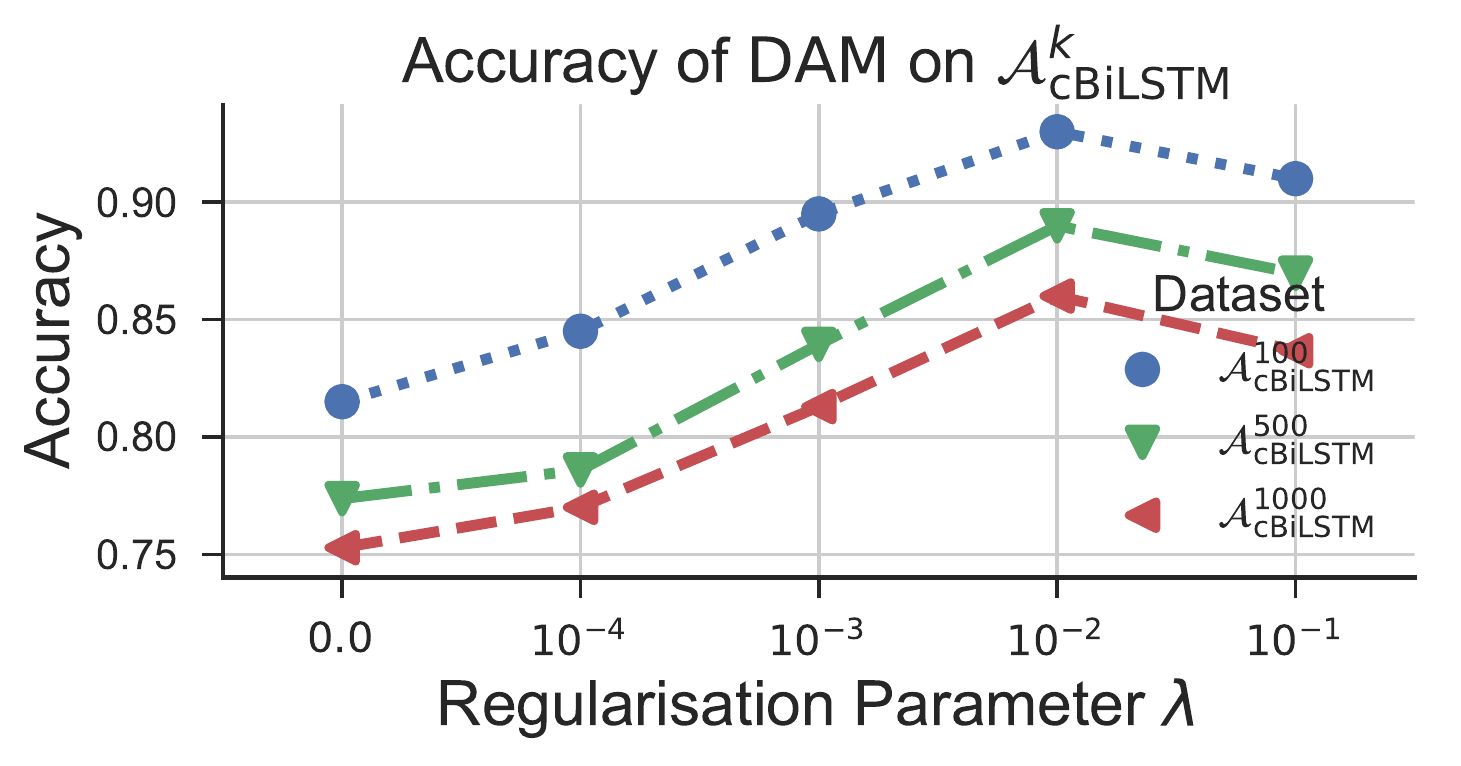}
In the following, we report the accuracy of \ESIM{} on several adversarial datasets $\aset{m}{k}$.
\includegraphics[width=\columnwidth]{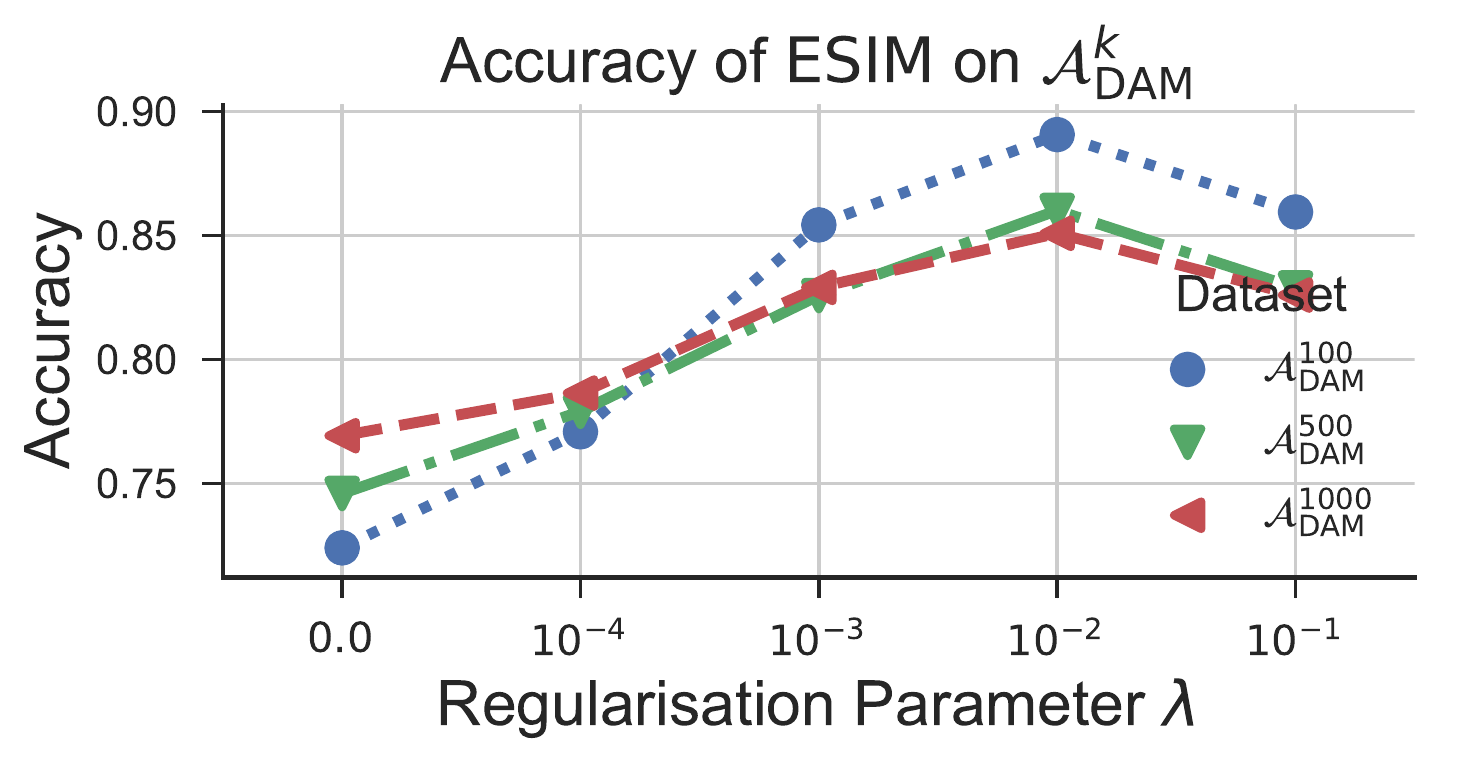}
\includegraphics[width=\columnwidth]{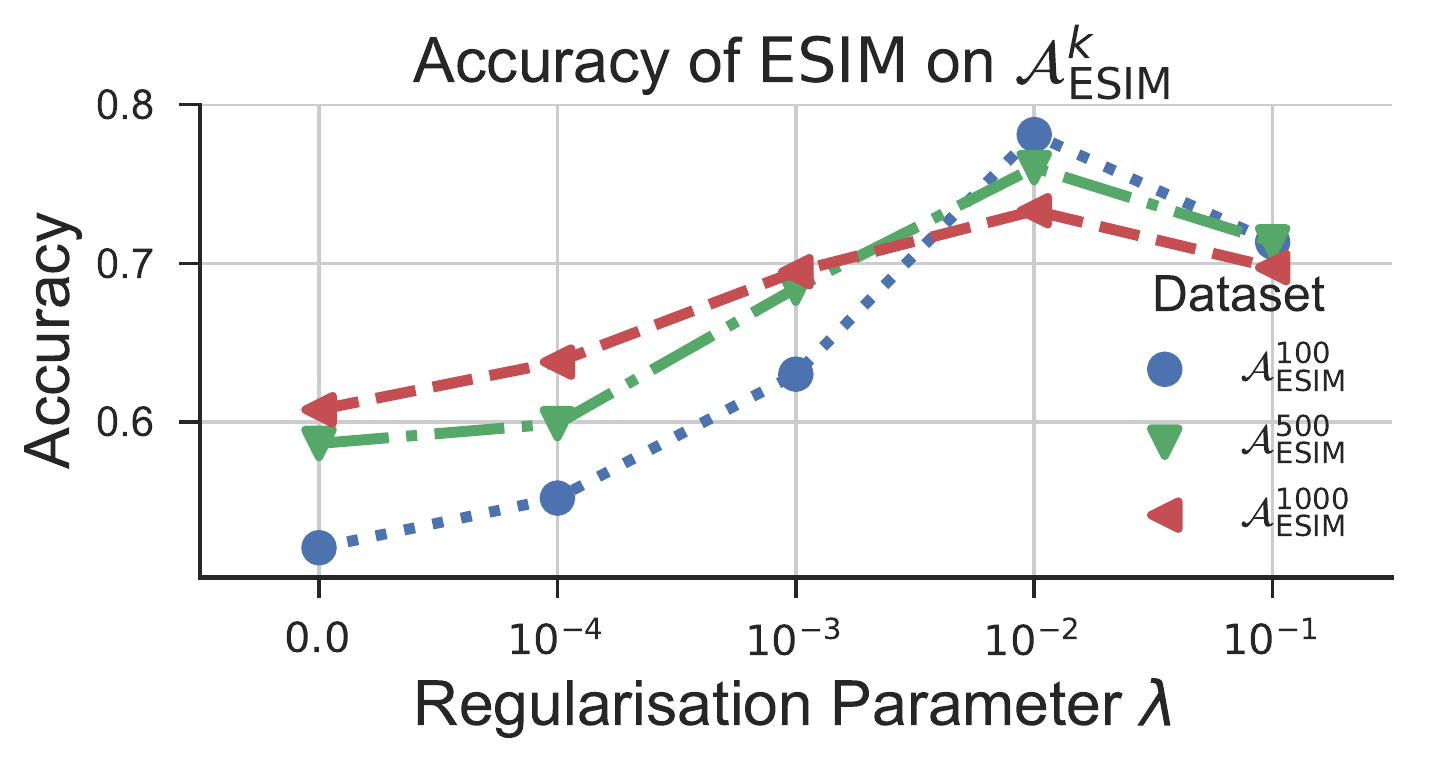}
\includegraphics[width=\columnwidth]{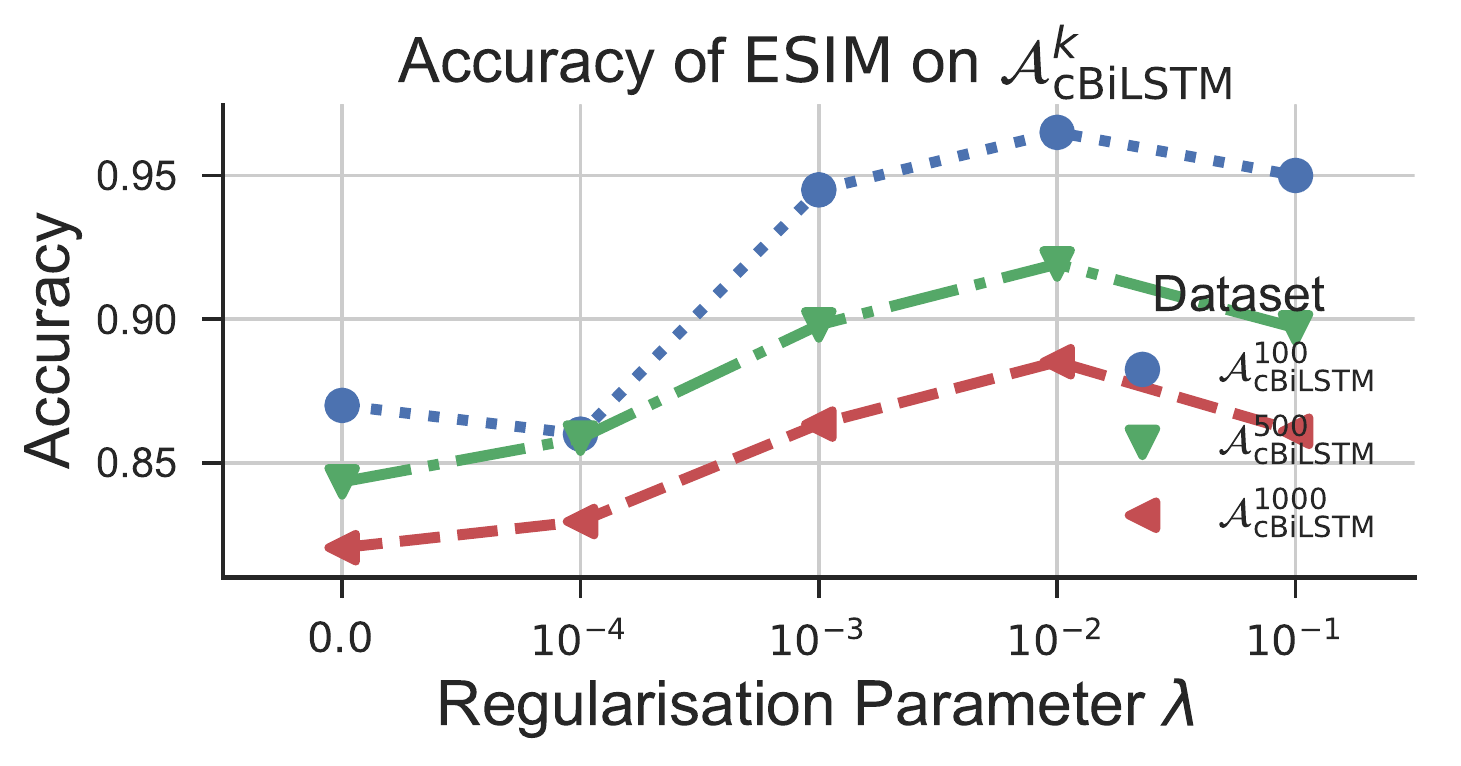}
In the following, we report the accuracy of \cBiLSTM{} on several adversarial datasets $\aset{m}{k}$.
\includegraphics[width=\columnwidth]{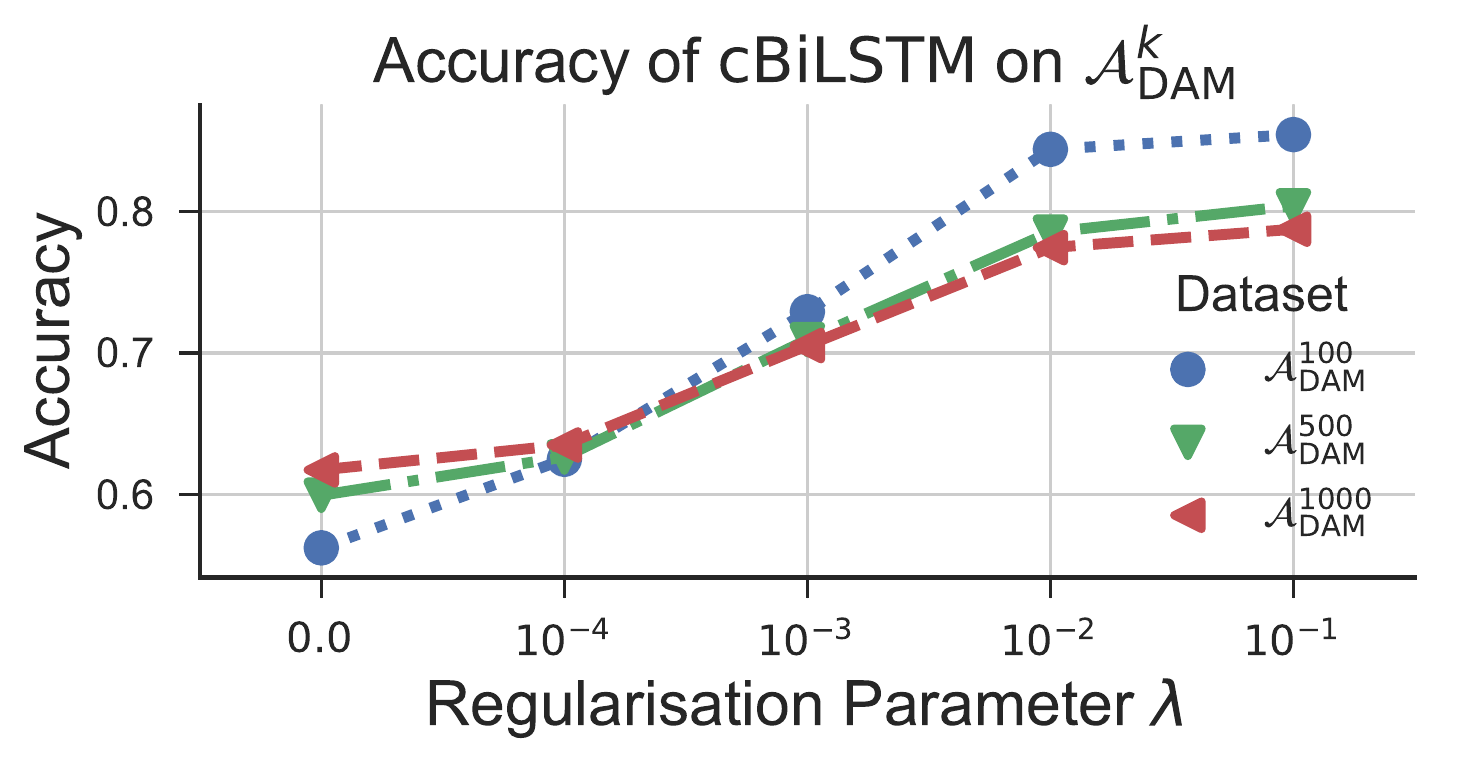}
\includegraphics[width=\columnwidth]{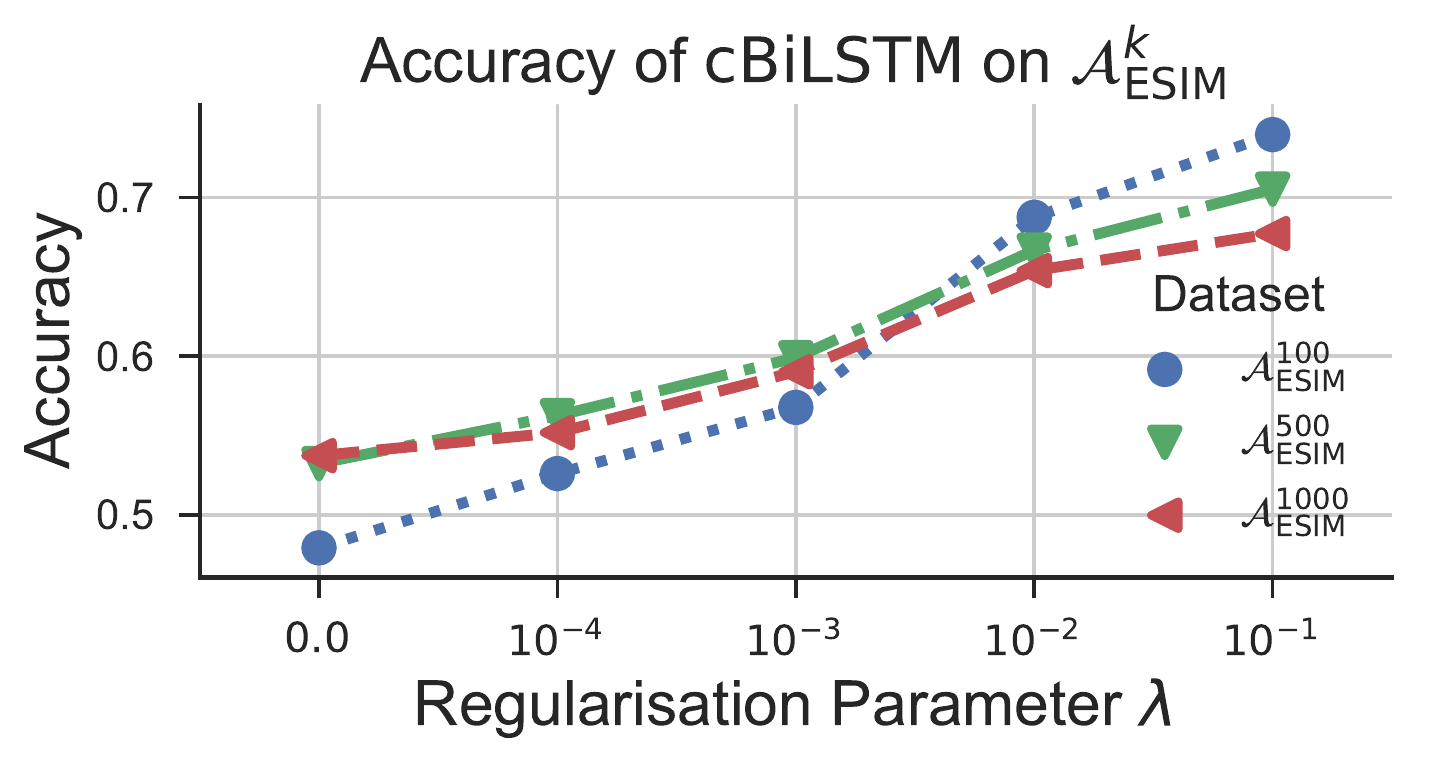}
\includegraphics[width=\columnwidth]{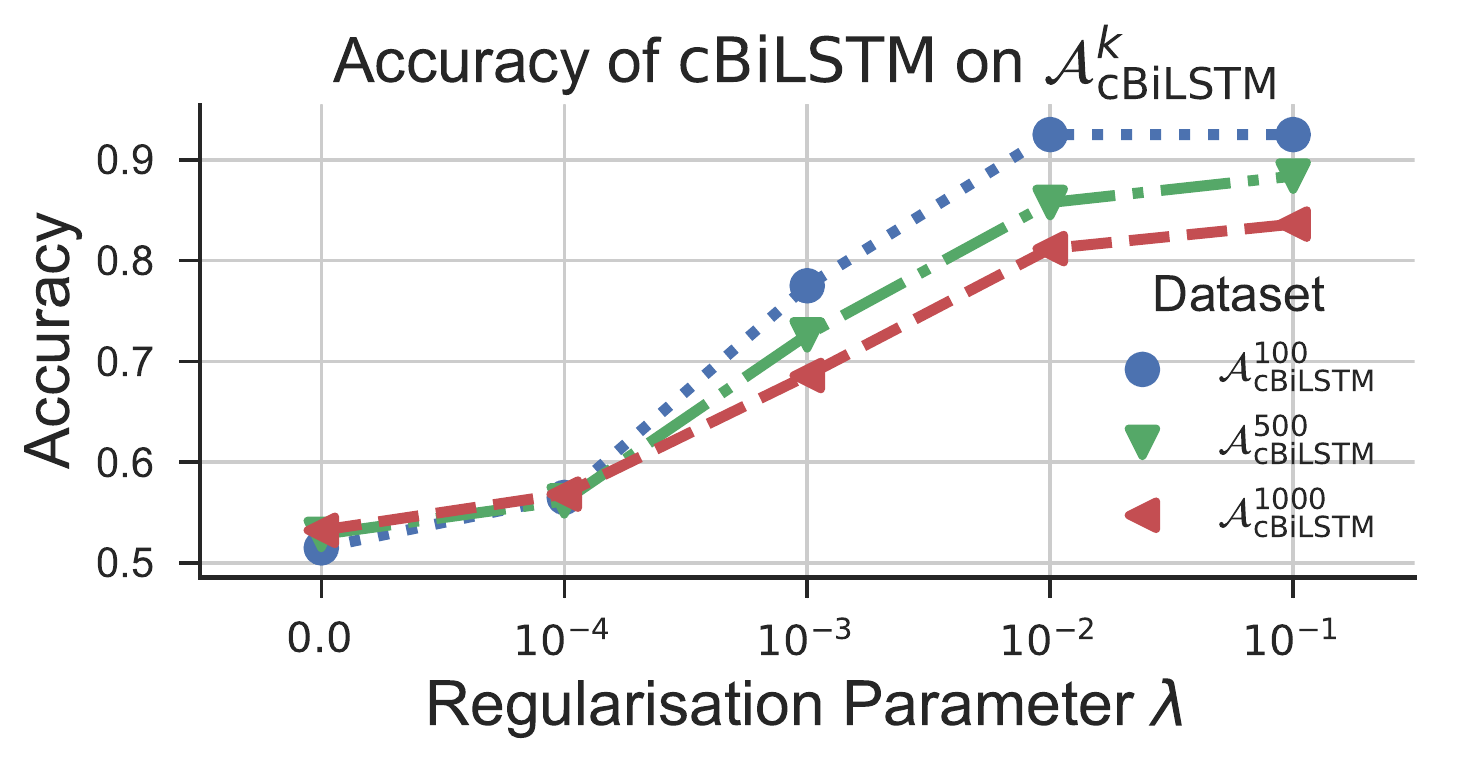}
\subsection{Adversarial examples} \label{sec:a:adversarial}
In \cref{tab:a:inconsistent} we report inconsistent results produced by \DAM{} on the SNLI training set, which violate rules $\xrule{2}$ and $\xrule{3}$ outlined in \cref{tab:rules}.
In \cref{tab:a:corruptions}, we report inconsistent results yield by \DAM{} on examples generated using the procedure described in \cref{ssec:discrete}.
\begin{table*}[ht]
 \centering
 \resizebox{\textwidth}{!}{
 \renewcommand{\arraystretch}{-1.0}
     \begin{tabular}{lrlc}
        \toprule
        & & {\bf Sentence} & {\bf Classification} \\
        \midrule
        \multirow{2}{*}{1} & $s_{1}$ & A young girl is holding a long thin yellow balloon. & $s_{1} \textcolor{red}{\xrightarrow{0.99}} s_{2}$ \\
        & $s_{2}$ & There is a girl watching a balloon & $s_{2} \textcolor{blue}{\xrightarrow{1.0}} s_{1}$ \\
        \midrule
        \multirow{2}{*}{2} & $s_{1}$ & A woman dressed in green is rollerskating outside at an event. & $s_{1} \textcolor{red}{\xrightarrow{1.0}} s_{2}$ \\
        & $s_{2}$ & A woman dressed in green is not rollerskating & $s_{2} \textcolor{blue}{\xrightarrow{0.76}} s_{1}$ \\
        \midrule
        \multirow{2}{*}{3} & $s_{1}$ & A young adult male, wearing black pants, a white shirt and a red belt, is practicing martial arts. & $s_{1} \textcolor{red}{\xrightarrow{0.99}} s_{2}$ \\
        & $s_{2}$ & A guy playing a video game on his flat screen television. & $s_{2} \textcolor{blue}{\xrightarrow{1.0}} s_{1}$ \\

        \midrule
        \midrule

        \multirow{2}{*}{4} & $s_{1}$ & Man sitting at a computer. & $s_{1} \textcolor{green}{\xrightarrow{1.0}} s_{2}$ \\
        & $s_{2}$ & The man is not outside running. & $s_{2} \textcolor{red}{\xrightarrow{1.0}} s_{1}$ \\
        \midrule
        \multirow{2}{*}{5} & $s_{1}$ & Two young women wearing bikini tops and denim shorts walk along side an orange VW Beetle. & $s_{1} \textcolor{green}{\xrightarrow{1.0}} s_{2}$ \\
        & $s_{2}$ & Two young women are not wearing coats and jeans & $s_{2} \textcolor{red}{\xrightarrow{1.0}} s_{1}$ \\
        \midrule
        \multirow{2}{*}{6} & $s_{1}$ & A woman in a hat sits reading and drinking a coffee. &  $s_{1} \textcolor{green}{\xrightarrow{1.0}} s_{2}$ \\
        & $s_{2}$ & martial arts demonstration & $s_{2} \textcolor{red}{\xrightarrow{1.0}} s_{1}$ \\
        \bottomrule
    \end{tabular}
 }
 \caption{
 Inconsistent results yield by \DAM{} on the SNLI training set.
 The notation $s_{1} \textcolor{red}{\xrightarrow{p}} s_{2}$ indicates that \DAM{} classifies the relation between $s_{1}$ and $s_{2}$ as \emph{contradiction} with probability $p$.
 We use different colours for representing the \textcolor{red}{contradiction}, \textcolor{green}{entailment} and \textcolor{blue}{neutral} classes.
 Examples $\{ 1, 2, 3\}$ (resp. $\{ 4, 5, 6 \}$) violate the logic rule $\xrule{2}$ (resp. $\xrule{3}$) in \cref{tab:rules}.
 }
\label{tab:a:inconsistent}
\end{table*}
\begin{table*}[ht]
    \centering
    \resizebox{\textwidth}{!}{
            \begin{tabular}{lrlc}
    \toprule
    & & {\bf Sentence} & {\bf Classification} \\
    \midrule
    \multirow{2}{*}{1} & $s_{1}$ & \makecell[l]{Two adults, one female in white, with shades and one male, gray clothes, walking across a street, away from a eatery with a \\ \quad \quad blurred image of a dark colored red shirted person in the foreground.} & $s_{1} \textcolor{red}{\xrightarrow{0.94}} s_{2}$ \\
    & $s_{2}$ & Two \sout{adults} \textcolor{red}{dogs} walk across a street. & $s_{2} \textcolor{green}{\xrightarrow{0.58}} s_{1}$ \\
    \midrule
    \multirow{2}{*}{2} & $s_{1}$ & A person on skis on a rail at night. & $s_{1} \textcolor{red}{\xrightarrow{0.89}} s_{2}$ \\
    & $s_{2}$ & They are \sout{fantastic} \textcolor{red}{sleeping} skiiers & $s_{2} \textcolor{blue}{\xrightarrow{0.66}} s_{1}$ \\
    \midrule
    \multirow{2}{*}{3} & $s_{1}$ & The school is having a special event in order to show the american culture on how other cultures are dealt with in parties. & $s_{1} \textcolor{red}{\xrightarrow{0.96}} s_{2}$ \\
    & $s_{2}$ & A \sout{school} \textcolor{red}{dog} is hosting an event. & $s_{2} \textcolor{blue}{\xrightarrow{0.66}} s_{1}$ \\
    
    \midrule
    \midrule
    
    \multirow{2}{*}{4} & $s_{1}$ & A woman is walking across the street eating a banana, while a man is following with his briefcase. & $s_{1} \textcolor{green}{\xrightarrow{0.96}} s_{2}$ \\
    & $s_{2}$ & A person that is \sout{hungry} \textcolor{red}{holding}. & $s_{2} \textcolor{red}{\xrightarrow{0.90}} s_{1}$ \\
    \midrule
    \multirow{2}{*}{5} & $s_{1}$ & A man and a woman cross the street in front of a pizza and gyro restaurant. & $s_{1} \textcolor{green}{\xrightarrow{0.95}} s_{2}$ \\
    & $s_{2}$ & Near a couple of \sout{restaurants} \textcolor{red}{picture}, two people walk across the street. & $s_{2} \textcolor{red}{\xrightarrow{0.82}} s_{1}$ \\
    \midrule
    \multirow{2}{*}{6} & $s_{1}$ & Woman in white in foreground and a man slightly behind walking with a sign for John's Pizza and Gyro in the background. & $s_{1} \textcolor{green}{\xrightarrow{0.98}} s_{2}$ \\
    & $s_{2}$ & The man with the sign is \sout{caucasian} \textcolor{red}{near}. & $s_{2} \textcolor{red}{\xrightarrow{0.51}} s_{1}$ \\
    
    \midrule
    \midrule
    
    
    \multirow{3}{*}{7} & $s_{1}$ & A boy is drinking out of a water fountain shaped like a woman. & $s_{1} \textcolor{green}{\xrightarrow{0.96}} s_{2}$ \\
    & $s_{2}$ & A male is getting a drink of water. & $s_{2} \textcolor{green}{\xrightarrow{0.93}} s_{3}$ \\
    & $s_{3}$ & A \sout{male} \textcolor{red}{man} is getting a drink of water. & $s_{1} \textcolor{red}{\xrightarrow{0.97}} s_{3}$ \\
    
    \midrule
    
    
    \multirow{3}{*}{8} & $s_{1}$ & A middle-aged oriental woman in a green headscarf and blue shirt is flashing a giant smile. & $s_{1} \textcolor{green}{\xrightarrow{0.98}} s_{2}$ \\
    & $s_{2}$ & A middle aged oriental woman in a green headscarf and blue shirt is flashing a giant smile & $s_{2} \textcolor{green}{\xrightarrow{0.97}} s_{3}$ \\
    & $s_{3}$ & A middle aged \sout{oriental} \textcolor{red}{young} woman in a green headscarf and blue shirt is flashing a giant smile & $s_{1} \textcolor{red}{\xrightarrow{0.67}} s_{3}$ \\
    
    \midrule
    
    
    \multirow{3}{*}{9} & $s_{1}$ & Bicyclists waiting at an intersection. & $s_{1} \textcolor{green}{\xrightarrow{0.96}} s_{2}$ \\
    & $s_{2}$ & The bicycles are on a road. & $s_{2} \textcolor{green}{\xrightarrow{0.94}} s_{3}$ \\
    & $s_{3}$ & \sout{The} \textcolor{red}{riding} bicycles are on a road. & $s_{1} \textcolor{red}{\xrightarrow{0.68}} s_{3}$ \\

    \bottomrule
    \end{tabular}
    }
    \caption{
    Inconsistent results produced by \DAM{} on adversarial examples generated using the discrete search procedure described in \cref{ssec:discrete} -- the pattern \sout{segment one} \textcolor{red}{segment two} denotes that the corruption process replaced ``segment one'' with ``segment two''.
    Examples $\{ 1, 2, 3 \}$ (resp. $\{ 4, 5, 6 \}$) violate the rule $\xrule{2}$ (resp. $\xrule{4}$), while examples $\{ 7, 8, 9 \}$ violate the logic rule in $\xrule{5}$.
    }
    \label{tab:a:corruptions}
\end{table*}

\clearpage

\subsection{Background Knowledge Violations} \label{sec:a:violations}
In the following we report the number of violations (\%) to rules in \cref{tab:rules} made by \DAM{}, \ESIM{}, and \cBiLSTM{} on the SNLI test set.
\bigbreak
\includegraphics[width=\columnwidth]{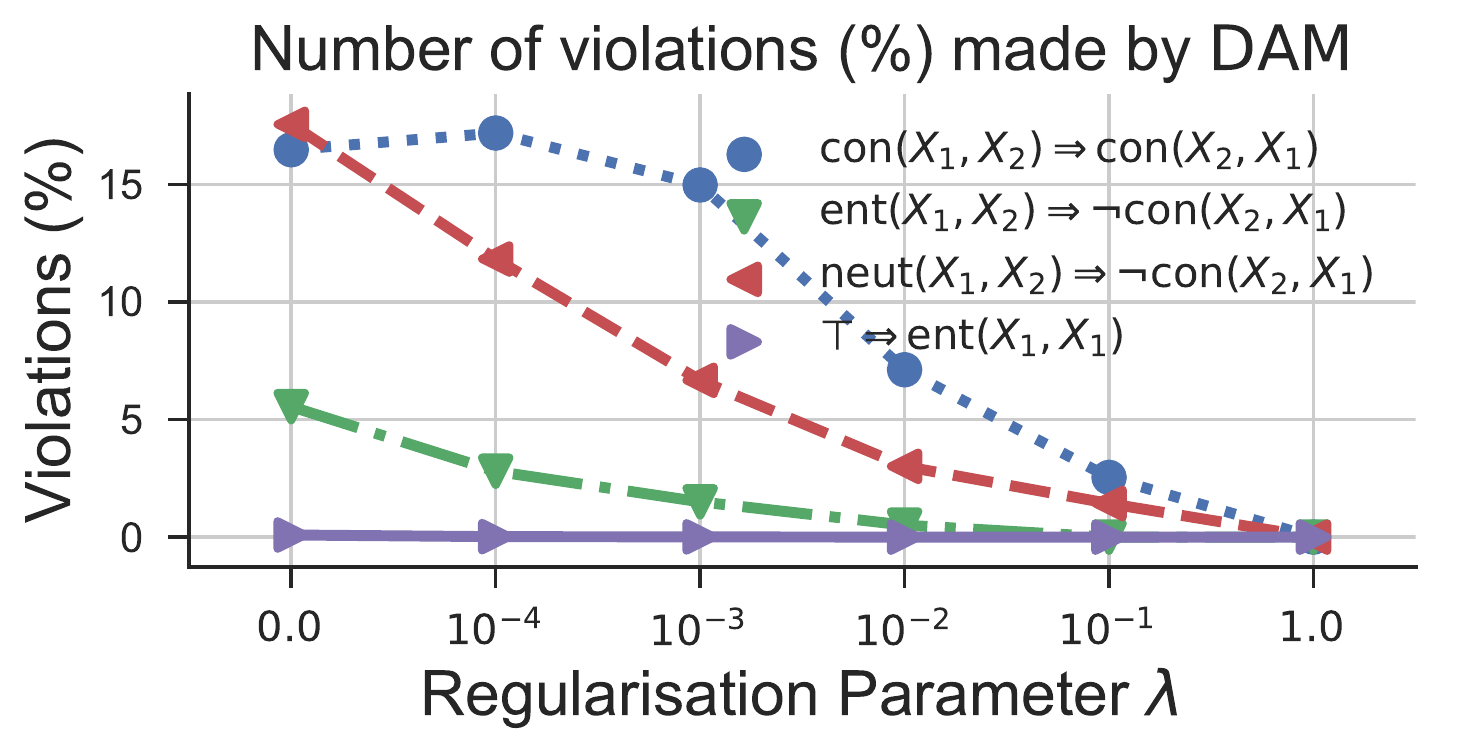}
\includegraphics[width=\columnwidth]{images/test_violations_esim_3_2.pdf}
\includegraphics[width=\columnwidth]{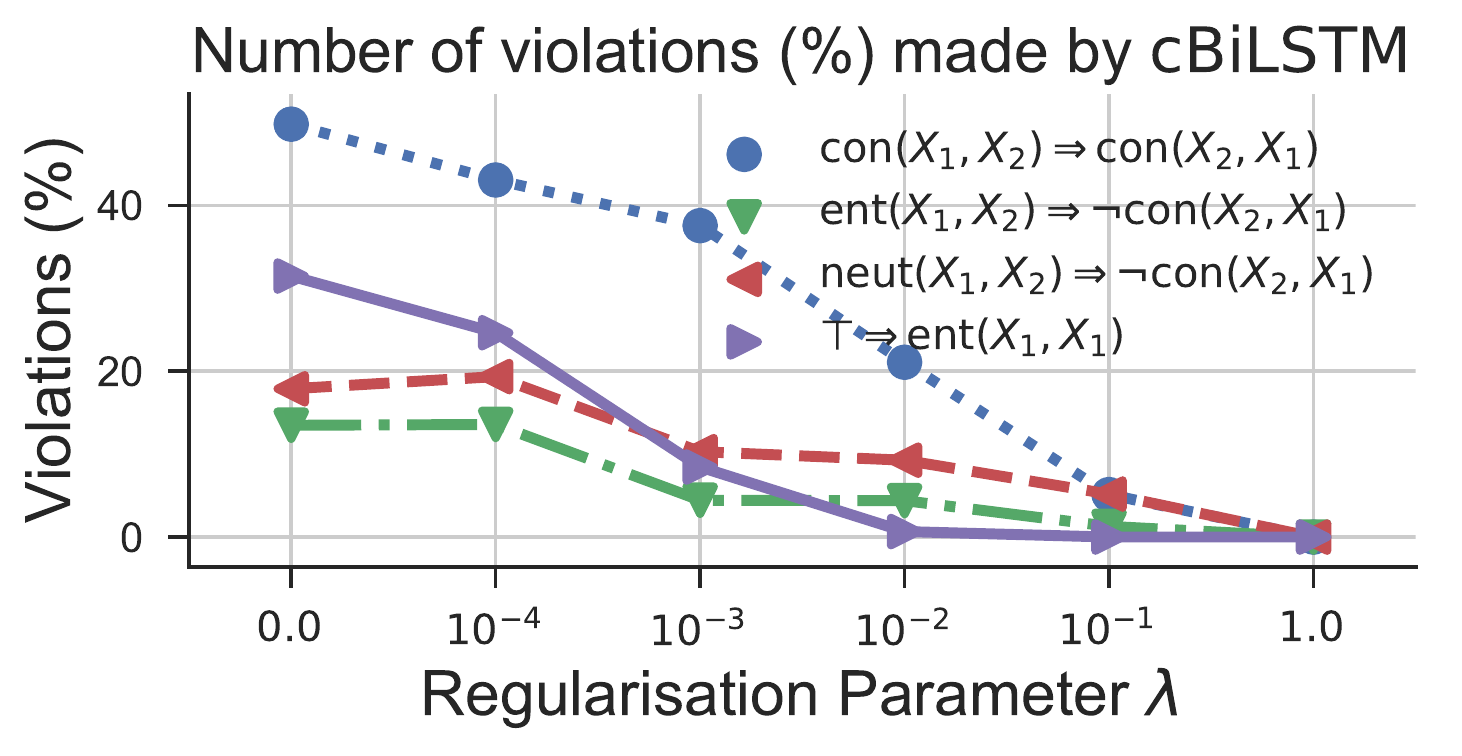}

\vfill\null

\subsection{Optimisation algorithms} \label{sec:a:optimisation}

In \cref{alg:a:dsearch} we describe our algorithm for generating adversarial examples by perturbing sentences in a dataset, and using a language model for constraining the generation process.
In \cref{alg:a:opt} we describe our adversarial training algorithm: it solves a minimax problem, where first a set of adversarial examples is generated by maximising the inconsistency loss $\iloss$.
Then, the model is trained by jointly minimising the data loss $\dloss$ and inconsistency loss on the generated adversarial examples.
\begin{algorithm}
\caption{Generation of Adversarial Sentences via Stochastic Perturbation Re-ranking}
\label{alg:a:dsearch}
\begin{algorithmic}
    \REQUIRE Perplexity threshold $\tau \in \Real_{+}$
    \STATE \COMMENT{Sample seed sentences from the dataset}
    \STATE $\sset \leftarrow \rel{sample}(\data)$
    \STATE \COMMENT{Generate a set of candidates, excluding the ones with a perplexity higher than $\tau$}
    \STATE $\mathcal{P} \leftarrow \{ \widetilde{\sset} \in \rel{perturb}(\sset) \mid  \log \langm(\widetilde{\sset}) \leq \tau \}$
    \STATE \COMMENT{Return the perturbations that maximise the inconsistency loss $\iloss$}
    \RETURN $\displaystyle{\arg\max_{\widetilde{\sset} \in \mathcal{P}} \iloss(\widetilde{\sset})}$
\end{algorithmic}
\end{algorithm}
\begin{algorithm}
 \caption{Solving the optimisation problem in \cref{eq:jloss} via Mini-Batch Gradient Descent} \label{alg:a:opt}
 \begin{algorithmic}[1]
 \REQUIRE{Dataset $\data$, weight $\lambda \in \Real_{+}$}
 \REQUIRE{No. of epochs $\tau \in \Natural_{+}$}
 \REQUIRE{No. of adv. substitution sets $n_{a} \in \Natural_{+}$}
  \STATE{\COMMENT{Initialise the model parameters $\hat{\params}$}}
  \STATE{$\hat{\params} \leftarrow \text{initialise}()$}
  \FOR{$i \in \{ 1, \ldots, \tau \}$}
   \FOR{$\data_{j} \in \text{batches}(\data)$}
    \STATE{\COMMENT{Generate the adv. substitution sets $S_{i}$}}
    \STATE{$\{ \sset_{1}, \ldots, \sset_{n_{a}} \} \leftarrow \text{generate}(\data_{j})$} \label{alg:a:opt:gen}
    \STATE{\COMMENT{Compute the gradient of \cref{eq:jloss}}}

    \STATE{$\mathcal{L} \leftarrow \dloss(\data_{j}, \hat{\params}) + \lambda \sum_{k = 1}^{n_{a}} \iloss(S_{k}; \hat{\params})$}
    \STATE{$g \leftarrow \nabla_{\params} \mathcal{L}$} \label{alg:a:opt:grad1}
    
    \STATE{\COMMENT{Update the model parameters}}
    \STATE{$\hat{\params} \leftarrow \hat{\params} - \eta g$} \label{alg:a:opt:update}
   \ENDFOR
  \ENDFOR
  \STATE{\textbf{return} $\hat{\params}$}
  \end{algorithmic}
\end{algorithm}
\end{document}